\documentclass{article}

\usepackage{amsfonts,amsopn,amsthm}
\usepackage{geometry,authblk}
\usepackage{graphicx}
\usepackage{algorithm,algpseudocode}
\usepackage{mathtools}
\usepackage{nicematrix,booktabs,tabularx,multirow,multicol}
\usepackage{cleveref,subcaption,enumitem}
\usepackage{siunitx}
\usepackage[normalem]{ulem}

\crefname{hypothesis}{Hypothesis}{Hypotheses}
\crefname{fact}{Fact}{Facts}
\crefname{section}{Section}{Sections}
\crefname{subsection}{Section}{Sections}
\Crefname{subsection}{Section}{Sections}
\DeclareMathOperator{\diag}{diag}
\numberwithin{equation}{section}
\newtheorem{theorem}{Theorem}

\newtheorem{lemma}[theorem]{Lemma}
\newtheorem{proposition}[theorem]{Proposition}
\graphicspath{{pics/}}
\title{Hybrid Least Squares\slash{}Gradient Descent Methods for DeepONets}
\author[1]{Jun Choi}
\author[1]{Chang-Ock Lee}
\author[2]{Minam Moon}
\affil[1]{\small Department of Mathematical Sciences, KAIST, Daejeon 34141, KOREA}
\affil[2]{\small Department of Mathematics, Korea Military Academy, Seoul 01805, KOREA}
\date{}

\begin{document}
\maketitle

\begin{abstract}
We propose an efficient hybrid least squares/gradient descent method to accelerate DeepONet
training. Since the output of DeepONet can be viewed as
linear with respect to the last layer parameters of the branch network,
these parameters can be optimized using a least squares~(LS) solve,
and the remaining hidden layer parameters are updated by means of gradient descent form.
However, building the LS system for all possible combinations
of branch and trunk inputs yields a prohibitively large linear problem that is infeasible to
solve directly. To address this issue, our method decomposes the large LS system into two
smaller, more manageable subproblems --- one for the branch network and one for the trunk
network --- and solves them separately.
This method is generalized to a broader type of $L^2$ loss with
a regularization term for the last layer parameters, including the case of unsupervised
learning with physics-informed loss.
\end{abstract}

\textbf{Key words.} hybrid least squares\slash{}gradient descent method, DeepONet,
physics-informed loss

\textbf{MSC codes.} 47-08, 65F45, 65Y10, 68T07, 68T20 

\let\thefootnote\relax\footnotetext{
    \textbf{Funding:} This work was supported by the National Research Foundation~(NRF) of
    Korea grant funded by the Korean government~(MSIT) [grant number RS2023--00208914].
}

\section{Introduction}\label{sec1}

The recent emergence of deep learning~(DL) has impacted the field of scientific computing,
which includes simulations of various dynamics and physical phenomena governed by partial
differential equations~(PDEs).
In particular, there has been a strong interest in replacing traditional numerical PDE solvers
with deep neural networks~(DNNs) so that the trained models can generate PDE solutions for
moderate unseen data, such as initial/boundary conditions~(IBCs), coefficients, or source
terms, etc.
To this end, Raissi et al.~\cite{Raissi2019} introduced Physics-Informed Neural 
Networks~(PINNs), where the solution of a PDE is expressed as a DNN, and the training of the 
DNN serves as the solving step of the PDE\@.
PINNs use the automatic differentiation technique~\cite{Baydin2018} in backpropagation for
the physics-informed loss~(PI-loss) with residual terms and IBC terms, with or without labeled
data. However, each PINN can handle only a single PDE instance at a time. If one of the PDE
components changes, further training is needed to solve the modified PDE\@.
This makes it difficult for PINNs to be fast surrogates for traditional numerical solvers.

To obtain DL solutions that vary with changes in the components of a PDE, a mapping between
these components and the solutions is needed.
In particular, these components can be functions, scalars, or vectors.
This has led to increased interest in neural operators, deep architectures designed
to learn mappings between infinite-dimensional function spaces
(e.g., mapping a coefficient function to a solution field).
Examples include Deep Operator Networks~(DeepONets)~\cite{Lu2021}, Fourier Neural
Operators~\cite{Li2020}, and Graph Kernel Networks~\cite{Li2020neural}.
These models aim to serve as fast surrogates for PDE solvers, albeit sometimes at the expense
of accuracy.
Notably, DeepONet is the most widely used neural operator since it has a theoretical
foundation in the Universal Approximation Theorem for operators~\cite{Chen1995} and offers
flexibility in input and output domains.
Furthermore, recent work demonstrated that physics-informed techniques
(i.e., including physical law terms in the loss) can be integrated into DeepONet
training~\cite{Wang2021}, thereby combining operator learning with physics-informed loss.
Additionally, DeepONet can be generalized to a multiple input operator network
(MIONet)~\cite{Jin2022}, which is a neural operator with multiple input functions.

Since DNN training generally requires a long time, various methods have been proposed to
accelerate training, including model compression via network pruning~\cite{Zheng2016},
better-conditioned optimization through weight normalization~\cite{Salimans2016},
accelerated convergence with methods like sparse momentum~\cite{Dettmers2019} or
hybrid least squares/gradient descent~(LSGD)~\cite{Cyr2020},
as well as domain-specific strategies such as multi-fidelity or parallel-in-time
training~\cite{Lee2022, Lee2024}.
For DeepONet, to optimize training, Lu et al.~\cite{Lu2022} eliminated computational redundancy
by using the same measurement locations for all output functions.
However, DeepONet’s structure — effectively the coupling of two neural networks via an inner
product — is more complex than a standard single-network model.
Consequently, the computational cost of training DeepONet is high.
This complexity makes it challenging to adopt existing techniques that accelerate DNN training
while achieving better convergence.

In this paper, we apply the hybrid LSGD method, which has shown successful results in
accelerating the training speed of DNNs, to vanilla DeepONets.
LSGD alternates between least squares~(LS) steps to find optimal coefficients
for fixed basis functions and gradient descent~(GD) steps to optimize the basis functions
which depend on the hidden layer parameters.
To this end, using the basis functions and coefficients for DeepONet, we construct LS problems
from given $l_2$ losses. However, since the LS system is too large to handle directly,
we develop a factorization technique that converts the large LS problem into
a special type of structured matrix equation with two smaller subproblems.
This allows for efficient solving and significant reduction of the computational burden.
Furthermore, we propose the least squares plus Adam~(LS+Adam) method as a
practical algorithm for LSGD, similar to applying the L-BFGS optimizer~\cite{Liu1989}
after Adam in PINN training.
Note that there is a similar work~\cite{Son2025}, Extreme Learning Machine
for DeepONet, which solves an LS problem by fixing randomly chosen hidden layer parameters.
Unlike~\cite{Son2025}, our method can be extended to DeepONet with a general type
of $L^2$ loss with a regularization term for the coefficients, which includes PI-loss for
linear PDEs\@. The regularization term ensures full-rank of the LS system.

The paper is organized as follows.
In \cref{sec2}, we provide preliminary understandings of the universal approximation
theorem for operators, its corresponding neural network, DeepONet.
We also explain the concept of LSGD in DNN training.
In \cref{sec3}, we formulate the LS problem from the sum of squared $l_2$ errors and
present LSGD methods for DeepONets.
In \cref{sec4}, we conduct experiments on various PDE problems to compare the training
performance between classical DeepONet training and DeepONet training with LS+Adam.

\section{Preliminaries}\label{sec2}
In this section, we introduce the neural operator DeepONet~\cite{Lu2021}
with its universal approximation theorem~\cite{Chen1995}
and the hybrid LSGD optimization method~\cite{Cyr2020}.
These form the theoretical foundation for our proposed training strategy described
in \cref{sec3}.

\subsection{Universal Approximation Theorem for Operator and DeepONet}\label{sec21}

Chen and Chen~\cite{Chen1995} showed that a nonlinear continuous operator can be
approximated by the inner product of a two-layer neural network and
a one-layer neural network with an appropriate activation function. The precise statement is
given as follows:

\begin{theorem}[Universal Approximation Theorem for Operator]\label{thm:UATO}
    Let $\sigma\colon \mathbb{R}\to\mathbb{R}$ be a Tauber-Wiener~(TW) function.
    That is, the set of all linear combinations of the form
    $\sum_{i=1}^{I} c_i \sigma(\lambda_i x+ \theta_i)$ with $\lambda_i,\theta_i,c_i\in\mathbb{R}$
    for $i=1,\dots,I$ is dense in every $C([a,b])$.
    Let $X$ be a Banach space, $K_1\subset X$, $K_2\subset\mathbb{R}^n$ be compact subsets
    of $X$ and $\mathbb{R}^n$, respectively, $V$ be a compact set in $C(K_1)$, and $G$ be a
    nonlinear continuous operator which maps $V$ into $C(K_2)$.
    Then, for any $\epsilon >0$, there exist positive integers $I,J,M$, real constants
    $c_{ij},\zeta_i,\theta_{ij}, \xi_{ij}^{m} \in \mathbb{R}$, vectors
    $\omega_i \in \mathbb{R}^n$, and $x_m\in K_1$
    with $i=1,\dots,I$, $j=1,\dots,J$, $m=1,\dots,M$ such that
    \begin{equation*}\label{eq:UATO}
        \left| G(u)(y) - \sum_{i=1}^{I}
        \underbrace{\sum_{j=1}^{J} c_{ij} \sigma\left(\sum_{m=1}^{M}
        \xi_{ij}^m u(x_m) + \theta_{ij}\right)}_{\textrm{two-layer network}} \cdot
        \underbrace{\sigma \left(\omega_i\cdot y + \zeta_i\right)}_{\textrm{one-layer network}}
        \right| < \epsilon
    \end{equation*}
    holds for all $u\in V$ and $y\in K_2$.
\end{theorem}
The conditions for the activation function $\sigma$ being a TW function are given
in~\cite{Barron1993,Chen1995,Cybenko1989,Hornik1989}.
For example, if $\sigma \in \mathcal{S}^{\prime} (\mathbb{R}) \cap C(\mathbb{R})$,
then $\sigma$ is a TW function if and only if $\sigma$ is not a polynomial.
Also, if $\sigma$ is a bounded sigmoid function such that $\lim\limits_{x\to -\infty}
\sigma(x) = 0$ and $\lim\limits_{x\to \infty} \sigma(x) = 1$, then $\sigma$ is a TW function.
Here, $\mathcal{S}^{\prime} (\mathbb{R})$ denotes the class of tempered distributions,
which is the collection of linear functionals defined on the Schwartz space
\begin{equation*}\label{Schwartz}
    \mathcal{S} (\mathbb{R})=\left\{\phi\in C^{\infty}(\mathbb{R}) ~\middle|~
    \sup\limits_{x\in\mathbb{R}} {\left|x^{\alpha}(D_x^{\beta}\phi)(x)\right|} < \infty
    \text{ for all nonnegative integers }\alpha,\beta\right\}.
\end{equation*}

Based on the above approximation theorem, Lu et al.~\cite{Lu2021} proposed a neural operator
called DeepONet, which generalizes the structure of the two networks.
The two-layer neural network that encodes input function values is called a branch network,
and the one-layer network that encodes output coordinates is called a trunk network.
Both branch and trunk networks can be deep neural networks of any structure, but the number
of output units must be the same. The corresponding universal approximation properties for
DeepONet were also provided in~\cite{Lu2021}:
\begin{theorem}[Generalized Universal Approximation Theorem for Operator]\label{thm:GUATO}
    Let $X$ be a Banach space, $K_1\subset X$, $K_2\subset\mathbb{R}^n$ be compact subsets
    of $X$ and $\mathbb{R}^n$, respectively, $V$ be a compact set in $C(K_1)$, and $G$ be a
    nonlinear continuous operator which maps $V$ into $C(K_2)$.
    Then, for any $\epsilon >0$, there exist positive integers $I,M$,
    continuous vector functions $\mathbf{b}\colon \mathbb{R}^M \to \mathbb{R}^I$ and
    $\mathbf{t}\colon \mathbb{R}^n \to \mathbb{R}^I$, and $x_m\in K_1$
    with $m=1,\dots,M$ such that
    \begin{equation*}\label{eq:GUATO}
        \left| G(u)(y) - \left\langle
        \underbrace{\mathbf{b}\left(u(x_1),\dots,u(x_M)\right)}_{\textrm{branch}},
        \underbrace{\mathbf{t}\left(y\right)}_{\textrm{trunk}}
        \right\rangle\right| < \epsilon
    \end{equation*}
    holds for all $u\in V$ and $y\in K_2$, where $\langle\cdot,\cdot\rangle$ denotes the
    inner product in $\mathbb{R}^I$.
    Moreover, the functions $\mathbf{b}$ and $\mathbf{t}$ can be chosen as diverse classes
    of neural networks satisfying the classical universal approximation theorem of functions.
\end{theorem}
\noindent{} For example, fully connected neural networks~\cite{Cybenko1989,Hornik1989,Lu2017},
convolutional neural networks~\cite{Heinecke2020,Zhou2020}, and
residual networks~\cite{Lin2018} have the universal approximation property.
We refer to~\cite{Augustine2024} and references therein for more details.

\begin{figure}[tb]
  \centering
  \includegraphics[width=0.5\linewidth]{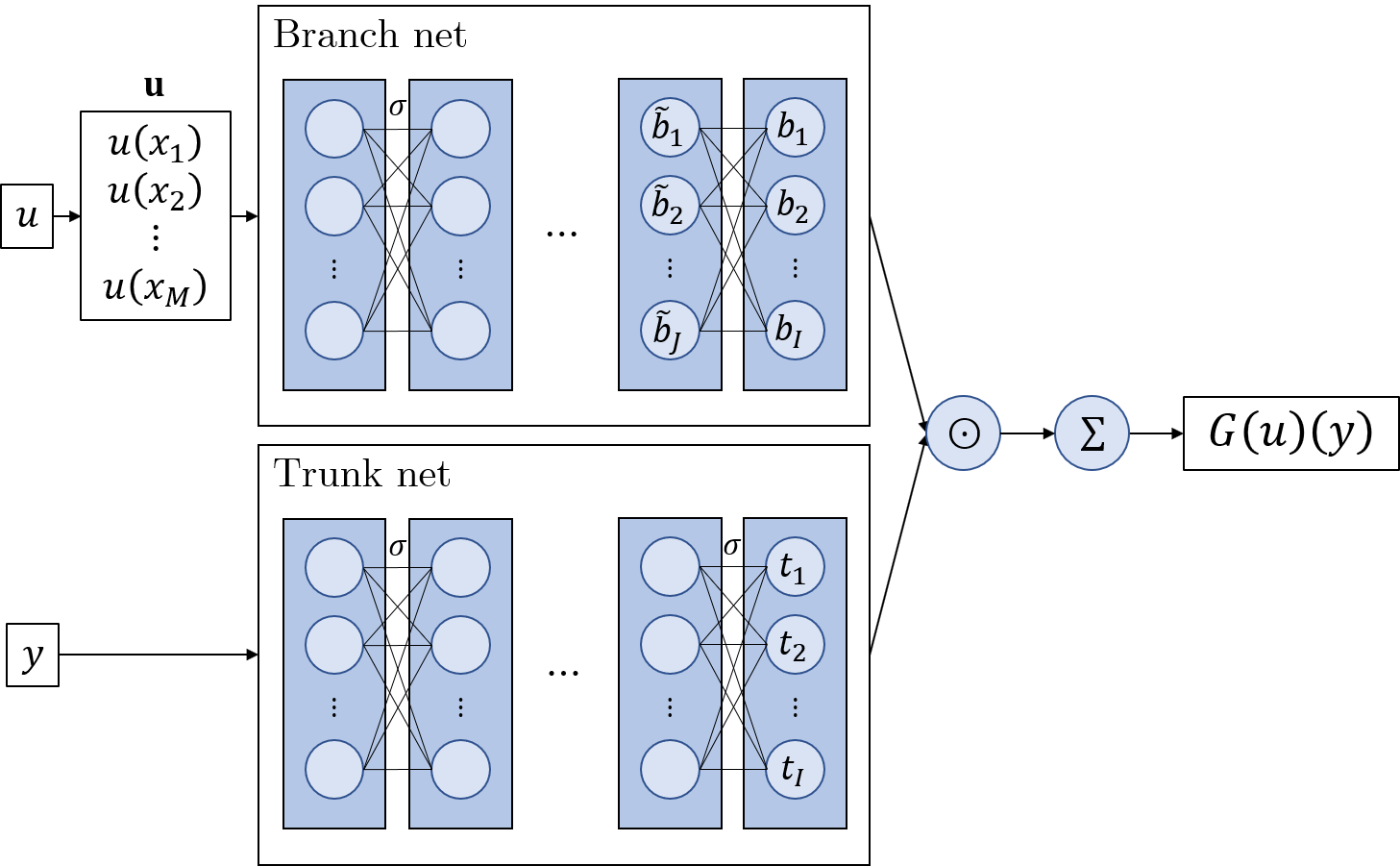}
  \caption{Structure of DeepONet with fully connected layers.}\label{fig:21}
\end{figure}

In this paper, we will focus on the structure where
the last layer of the branch network is a fully connected layer without bias and
activation function.
This structure corresponds to the network in \cref{thm:UATO} and possesses the
universal approximation property. See \cref{fig:21}, where $\odot$ denotes the Hadamard
(entrywise) product of vectors with the same length.

Let $G$ be an operator which takes an input function $u$, then the corresponding output function
is $G(u)$.
For any $y\in{\mathbb{R}}^d$ in the domain of $G(u)$, let $G(u)(y)\in\mathbb{R}$.
The input function $u$ is discretized as
$\mathbf{u}={[u(x_1),\dots, u(x_M)]}^{T}\in{\mathbb{R}}^M$,
where the input sensors ${\{x_m\}}_{m=1}^{M}$ are fixed for the input function $u$.
Here, the superscript ${T}$ denotes the transpose of a matrix or a vector
except $\theta^{T}$ in \cref{sec3}.
The branch network maps $\mathbf{u}$ to $\mathbf{b}(\mathbf{u}) =
{[b_1(\mathbf{u}),\dots,b_I(\mathbf{u})]}^{T} \in{\mathbb{R}}^I$.
Also, we denote the output immediately before the last layer as
$\tilde{\mathbf{b}}(\mathbf{u}) =
{[\tilde{b}_1(\mathbf{u}),\dots,\tilde{b}_J(\mathbf{u})]}^{T} \in{\mathbb{R}}^J$.
On the other hand, the trunk network maps $y\in{\mathbb{R}}^d$ to
$\mathbf{t}(y) = {[t_1(y),\dots,t_I(y)]}^{T} \in{\mathbb{R}}^I$.
Here, the DeepONet approximating $G$ is expressed as
\begin{equation} \label{eq:DON}
    G(u)(y)\approx \langle \mathbf{b}(\mathbf{u}), \mathbf{t}(y)\rangle
    = \sum_{i=1}^{I}b_i(\mathbf{u})t_i(y)
    = \sum_{i=1}^{I}\left(\sum_{j=1}^{J} c_{ij}\tilde{b}_j(\mathbf{u})\right)t_i(y),
\end{equation}
where $C = \left(c_{ij}\right)\in{\mathbb{R}}^{I\times J}$ is the parameter matrix of the
last layer of the branch net which maps ${\mathbb{R}}^J$ to ${\mathbb{R}}^I$.

\subsection{Hybrid Least Squares/Gradient Descent Method for Neural Networks}\label{sec23}
To improve the training of DNNs, Cyr et al.~\cite{Cyr2020} suggested the
hybrid LSGD method, which alternates between LS steps for the fully connected last layer
parameters and GD steps for other parameters.
Without loss of generality, we consider a neural network
$\text{NN}_\theta\colon\mathbb{R}^M\to\mathbb{R}$ with parameter $\theta$,
whose output layer is a linear layer without bias;
if there is a bias term, it can be considered as the product of the constant function
$\phi_0(x)\equiv 1$ and the bias coefficient.
Then, the output of the neural network is expressed as
\begin{equation}\label{eq:NNBasis}
    \text{NN}_{\theta}(\mathbf{x}) = \sum_{j=1}^{J} \theta_{j}^{L} \phi_j(\mathbf{x};\theta^H),
    \qquad \theta=(\theta^H,\theta^L),
\end{equation}
where $\theta^{L}=(\theta_j^L)$ is the output layer parameter, $\theta^H$ is other hidden
layer parameters, and $\phi_j(\mathbf{x})$ is the $j$-th input unit for the output layer.
Note that this can be viewed in terms of a linear combination of basis functions $\phi_j$
with coefficients $\theta_{j}^{L}$, and is the motif for the development of the LSGD algorithm.

Suppose that we solve a standard $l_2$ regression problem:
\begin{equation}\label{eq:minL2}
    \min_{\theta} {\left\|u-\text{NN}_{\theta}\right\|}_{l_2(\chi)}^{2}
\end{equation}
where $\chi={\{\mathbf{x}_p\}}_{p=1}^{P}$ is a finite collection of input data
and $u$ is the given target data.

A common way to minimize the loss \cref{eq:minL2} is to use a gradient descent type optimizer
for the whole parameter $\theta$. Alternatively, we can fix $\theta^H$ and minimize the loss
with respect to $\theta^L$. This gives an LS problem in terms of $\theta^L$:
\begin{equation*}\label{eq:minL2LS}
    \min_{\theta^L} {\left\|f-A\theta^L\right\|}_{2}^{2}
\end{equation*}
where $f=\left(u(\mathbf{x}_p)\right)\in\mathbb{R}^P$ and $A=\left(\phi_j(\mathbf{x}_p;\theta^B)
\right)\in\mathbb{R}^{P\times J}$. This is the LS step of the LSGD method.
For the GD step, fix $\theta^{L}$ and update $\theta^{H}$ using a GD-type optimizer such as GD,
stochastic GD~(SGD), or Adam~\cite{Kingma2017}.
The whole process is described in~\cite[Algorithm~1]{Cyr2020}.

Note that the LSGD method for \cref{eq:minL2} can be extended to more general losses that
consist of the sum of $K$ squared $l_2$ error terms with linear operators $\mathcal{L}_k$:
\begin{equation*}\label{eq:minGenL2LS}
    \sum_{k=1}^{K} \epsilon_{k} {\left\|\mathcal{L}_k\left[u\right] -
    \mathcal{L}_k\left[\text{NN}_\theta \right]\right\|}_{l_2(\chi_k)}^{2}
\end{equation*}
where $\chi_k={\{\mathbf{x}_{p_k}\}}_{p_k=1}^{P_k}$ is a finite collection of input data,
$\mathcal{L}_k\left[u\right]$ is the given target data,
and $\epsilon_k > 0$ is the weight for the $k$-th term.
These losses encompass the supervised learning with the standard $l_2$ loss ($K=1$ and
$\mathcal{L}$ being the identity) and the unsupervised learning for linear PDE with PI-loss
($K>1$ and $\mathcal{L}_k$ being either the residual operator for the PDE or the
initial/boundary condition for the domain).

\section{Hybrid Least Squares/Gradient Descent Method for DeepONets}\label{sec3}
In this section, we formulate the hybrid LSGD training schemes for operator learning.
Consider the following loss for DeepONet, which consists of the sum of squared $l_2$
error terms indexed by $k$ and regularization for the last layer parameter of the branch network:
\begin{equation}\label{eq:GenL2LSDON}
    \sum_{k=1}^{K} \epsilon_{k} {\left\|\mathcal{L}_k\left[G(\cdot)\right](u,y) -
    \mathcal{L}_k\left[\langle \mathbf{b}(\cdot;\theta^{B},\theta^{L}),
    \mathbf{t}(\cdot;\theta^{T})\rangle\right](\mathbf{u},y)\right\|}_{l_2(\chi_k)}^{2}
    + \lambda {\left\|\theta^{L}\right\|}_{2}^{2},
\end{equation}
where $\chi_k$ is a finite collection of data pairs $(u,y)$,
$\mathcal{L}_k$ is a linear operator between real-valued functions for the $k$-th error term,
$\epsilon_k > 0$ and $\lambda\geq 0$ are the weights for each error term and the regularization
term, and $\theta^{B}$, $\theta^{T}$, $\theta^{L}$
denote the parameters for the branch network except the last layer, the trunk network,
and the last layer of the branch, respectively.
If there is no confusion, we will refer to the last layer of the branch network of a DeepONet
as the last layer of the DeepONet.

Here, we represent the last layer parameter as
\begin{equation}\label{eq:vecCDON}
    \theta^{L}=\text{vec}({C^{T}})=
    {[c_{11},\dots,c_{1J},c_{21},\dots,c_{2J},\dots,c_{I1},\dots,c_{IJ}]}^{T}
    \in{\mathbb{R}}^{IJ},
\end{equation}
where $C=(c_{ij})\in\mathbb{R}^{I\times J}$ is the parameter matrix for the last layer and
$\text{vec}(X)$ denotes the column-wise vectorization of $X\in\mathbb{R}^{R\times S}$, i.e.,
\begin{equation*}\label{eq:vec}
    \text{vec}(X) = \begin{bmatrix}
        \mathbf{x}_1 \\
        \vdots \\
        \mathbf{x}_S
    \end{bmatrix} \in {\mathbb{R}}^{RS},
\end{equation*}
for $X = [\mathbf{x}_1\, \cdots\, \mathbf{x}_S]$ with $\mathbf{x}_s \in{\mathbb{R}}^{R}$.
Note that $\theta^{L}$ is the row-wise vectorization of $C$.

Instead of using a GD-type optimizer for all parameters
$(\theta^{B},\theta^{T},\theta^{L})$, we can utilize the LSGD method directly since
the DeepONet structure in \cref{eq:DON} corresponds to the representation
in \cref{eq:NNBasis} with basis functions $\tilde{b}_j t_i$ and coefficients $c_{ij}$.
The corresponding LS problem for the last layer parameter is
\begin{equation}\label{eq:LLLSDON}
    \min_{\theta^{L}} {\sum_{k=1}^{K} \epsilon_{k} {\left\|f_k - A_k \theta^{L}\right\|}_{2}^{2}
    + \lambda {\left\|{\theta}^L\right\|}_{2}^{2}},
\end{equation}
where $\chi_k={\{(\hat{u}_{d_k},\hat{y}_{d_k})\}}_{d_k=1}^{D_k}$ is the collection of data
pairs on which the loss is evaluated,
$f_k = \left({\mathcal{L}_k\left[G(\cdot)\right](\hat{u}_{d_k},\hat{y}_{d_k})}\right)
\in{\mathbb{R}}^{D_k}$ is the given data, and $A_k \in{\mathbb{R}}^{D_k\times IJ}$ is the
matrix whose $(d_k,J(i-1)+j)$ entry is
$\mathcal{L}_k\left[\tilde{b}_j t_i\right](\hat{\mathbf{u}}_{d_k},\hat{y}_{d_k})$ for
$1\leq d_k \leq D_k$, $1\leq i \leq I$, and $1\leq j \leq J$.

However, if $I$ and $J$ are not small enough, the size of the LS system \cref{eq:LLLSDON}
becomes too large to handle directly, unless the data size of the LS step, $D_k$, is small.
On the other hand, using small $D_k$ is more likely to lead to overfitting, and the
data not used in the training is more likely to cause larger errors.
This makes it difficult to apply the LSGD method directly to DeepONet training.

Suppose that the data collection $\chi_k$ can be represented as a Cartesian product of only
data $u$ and $y$, such as
\begin{equation}\label{eq:CondDataDON}
    \chi_k = \beta \times \tau_k,
\end{equation}
where $\beta = {\left\{u_p\right\}}_{p=1}^{P}$ is the set of input functions and
$\tau_k = {\left\{y_{q_k}\right\}}_{q_k=1}^{Q_k}$
is the set of points of the discretized domain for $\mathcal{L}_k [G(\cdot)]$ with
$D_k = P Q_k$.
Therefore, all input functions are used equally in all error terms, and
the same discretization for the domain of $\mathcal{L}_k [G(\cdot)]$
is applied to all input functions $u_p$ in each error term.

We also assume that the linear operator $\mathcal{L}_k$ satisfies
\begin{equation}\label{eq:CondOpDON}
    \mathcal{L}_k\left[\tilde{b}_j t_i\right](\mathbf{u},y)
    = \tilde{b}_j(\mathbf{u}) \mathcal{L}_k\left[t_i\right](y)
\end{equation}
for all $k$.
This implies the linear operator $\mathcal{L}_k$ is independent of the input function and
acts on each trunk component $t_i$.

Let $\otimes$ denote the Kronecker product, and let
the permutation matrix $K_{R,S}\in\mathbb{R}^{RS\times RS}$ be the commutation
matrix~\cite{Magnus1979}, which satisfies $\text{vec}(X^{T}) = K_{R,S} \text{vec}(X)$
for every $X\in\mathbb{R}^{R\times S}$. This permutes the column-wise vectorization of $X$ into
the row-wise vectorization of $X$ and satisfies $K_{R,S}^{T} = K_{S,R}$.
More specifically, $K_{R,S}$ can be understood as an $R\times S$ block matrix whose $(r,s)$
block $K_{R,S}^{(r,s)}$ is the $S \times R$ matrix whose entries are all zero except the
$(s,r)$ entry, which has the value one. That is,
\begin{equation*}\label{eq:CommBlk1}
    K_{R,S} = \begin{bmatrix}
        K_{R,S}^{(1,1)} & K_{R,S}^{(1,2)} & \cdots & K_{R,S}^{(1,S)}\\
        K_{R,S}^{(2,1)} & K_{R,S}^{(2,2)} & \cdots & K_{R,S}^{(2,S)}\\
        \vdots & \vdots & \ddots & \vdots\\
        K_{R,S}^{(R,1)} & K_{R,S}^{(R,2)} & \cdots & K_{R,S}^{(R,S)}
    \end{bmatrix},
\end{equation*}
where
\begin{equation*}\label{eq:CommBlk2}
    K_{R,S}^{(r,s)} = \begin{bNiceMatrix}[last-row,last-col]
        0 & \cdots & 0 & \cdots & 0 & \\
        \vdots & \ddots & \vdots & \ddots & \vdots & \\
        0 & \cdots & 1 & \cdots & 0 & \leftarrow s\\
        \vdots & \ddots & \vdots & \ddots & \vdots & \\
        0 & \cdots & 0 & \cdots & 0 & \\
        & & {\begin{matrix}\uparrow\\r\end{matrix}} & & &
    \end{bNiceMatrix}.
\end{equation*}

We will show that under the conditions \cref{eq:CondDataDON} and \cref{eq:CondOpDON},
the large matrix $A_k$ in \cref{eq:LLLSDON} can be factored into the
product of two smaller matrices with permutations.
This allows the LS problem to be reduced to a matrix equation of small size, which can be
solved by elementary methods.
\begin{theorem}\label{thm:DONFact}
    Under the conditions \cref{eq:CondDataDON} and \cref{eq:CondOpDON}, the large matrix $A_k\in
    \mathbb{R}^{PQ_k \times IJ}$ in the least squares problem \cref{eq:LLLSDON} can be factored as
    \begin{equation}\label{eq:DONFact}
        A_k = K_{P,Q_k} (T_k \otimes B),
    \end{equation}
    where $B = \left({\tilde{b}_j({\mathbf{u}}_p)}\right) \in {\mathbb{R}}^{P\times J}$ is
    the branch pre-output matrix,
    $T_k = \left({\mathcal{L}_k\left[t_i\right]({y}_{q_k})}\right) \in {\mathbb{R}
    }^{Q_k\times I}$ is the trunk output matrix with operator $\mathcal{L}_k$,
    and $K_{P,Q_k}\in{\mathbb{R}}^{PQ_k\times PQ_k}$ is the commutation matrix of $P\times Q_k$
    matrices.
\end{theorem}

Before proving the theorem, we denote the lexicographic order of the entry of and
$N$-dimensional tensor of size $D_1\times\cdots\times D_N$ as
\begin{equation}\label{eq:LexOrder}
    {[\alpha_1,\dots,\alpha_N]}_{D_1,\dots,D_N} \coloneq
    \sum\limits_{m=1}^{N-1} \left[\left(\prod\limits_{l=m+1}^{N} D_l \right)(\alpha_m - 1)
    \right] + \alpha_N,
\end{equation}
where $1\leq\alpha_m\leq D_m$ for each $m=1,\dots,N$.

Note that the commutation matrix $K_{D_1,D_2}$ satisfies
\begin{equation}\label{eq:CommStdBasis}
    \mathbf{e}_{{[\alpha_1,\alpha_2]}_{D_1,D_2}} = K_{D_1,D_2}
    \mathbf{e}_{{[\alpha_2,\alpha_1]}_{D_2,D_1}}
\end{equation}
where $\mathbf{e}_i$ is the $i$-th standard basis in $\mathbb{R}^{D_1D_2}$ as a column vector.
Also, the entry of the Kronecker product of two matrices $X\in\mathbb{R}^{R_1\times S_1}$ and
$Y\in\mathbb{R}^{R_2\times S_2}$ can be represented as
\begin{equation}\label{eq:KronNota}
    {(X\otimes Y)}_{r s} = {X}_{r_1 s_1} {Y}_{r_2 s_2},
\end{equation}
where $r = {[r_1,r_2]}_{R_1,R_2}$ and $s = {[s_1,s_2]}_{S_1,S_2}$.

\begin{proof}[Proof of \cref{thm:DONFact}]
    By the formulation of the LS problem and the lexicographic ordering of the last layer
    parameters \cref{eq:vecCDON}, $(d_k,{[i,j]}_{I,J})$ entry of $A_k$ is
    \begin{equation*}\label{eq:DONFactEntry}
        \mathcal{L}_k\left[\tilde{b}_j t_i\right](\hat{\mathbf{u}}_{d_k},\hat{y}_{d_k}).
    \end{equation*}
    By \cref{eq:CondDataDON}, for each $d_k$, there exist $p$ and $q$ such that
    $d_k = {[p,q]}_{P,Q_k}$, where $\hat{u}_{d_k} = {u}_p \in \beta$ and
    $\hat{y}_{d_k} = {y}_{q}\in\tau_k$. Also, the pair $(p,q)$ and $d_k$ have one-to-one
    correspondence as $1\leq p \leq P$, $1\leq q \leq Q_k$ and $1\leq d_k \leq D_k = PQ_k$.
    Hence, by \cref{eq:CondOpDON}, $({[p,q]}_{P,Q_k},{[i,j]}_{I,J})$ entry of $A_k$ can be
    expressed as
    \begin{equation}\label{eq:DONFactExpress}
        \tilde{b}_j({\mathbf{u}}_p) \mathcal{L}_k\left[t_i\right]({y}_{q}).
    \end{equation}

    On the other hand, by \cref{eq:KronNota}, $({[q,p]}_{Q_k,P},{[i,j]}_{I,J})$ entry of
    $T_k \otimes B$ is exactly \cref{eq:DONFactExpress}. By using \cref{eq:CommStdBasis}
    to rearrange the row order of $T_k \otimes B$, we have $A_k = K_{P,Q_k} (T_k \otimes B)$.
\end{proof}

Now, let us observe the LS problem \cref{eq:LLLSDON} using the result of \cref{thm:DONFact}.
If we express the given data vector $f_k$ in the vectorized form of a matrix, we have
\begin{equation*}\label{eq:vecFDON}
    f_k = \text{vec}(F_k^{T}),
\end{equation*}
where $F_k = \left(\mathcal[G(\cdot)]({u}_p,{y}_{q_k})\right)\in
\mathbb{R}^{P\times Q_k}$ is the matrix form of the given data. Therefore, the LS problem with
respect to the last layer parameter is
\begin{equation}\label{eq:LLLSDONFinal}
    \min_{C} {\sum_{k=1}^{K} \epsilon_{k} {\left\|\text{vec}(F_k^{T}) - K_{P,Q_k}
    (T_k \otimes B)
    \text{vec}(C^{T})\right\|}_{2}^{2} +
    \lambda {\left\|\text{vec}(C^{T})\right\|}_{2}^{2}}.
\end{equation}

Note that for the LS problems with multiple error terms, the minimizer is the solution of the
normal equation.
\begin{lemma}\label{lemma:LSNormal}
    For the least squares problem with positive weights $\epsilon_k > 0$,
    \begin{equation*}\label{eq:LSNormal1}
        \min_{\mathbf{x}} {\sum_{k=1}^{K} \epsilon_{k} {\left\|\mathbf{b}_k - A_k\mathbf{x}
        \right\|}_{2}^{2}},
    \end{equation*}
    the minimizer $\hat{\mathbf{x}}$ satisfies the normal equation:
    \begin{equation*}\label{eq:LSNormal2}
        \left(\sum_{k=1}^{K} \epsilon_k A_k^{T}A_k \right) \hat{\mathbf{x}}
        = \sum_{k=1}^{K} \epsilon_k A_k^{T}\mathbf{b}_k.
    \end{equation*}
\end{lemma}
Additionally, we recall following basic properties for the Kronecker
product~\cite{Magnus1979,Neudecker1968}:
\begin{lemma}\label{lemma:KronProp}
    \begin{enumerate}[label={(\roman*)}]
        \item $(X \otimes Y)(Z \otimes W) = (X Z)\otimes(Y W)$,
        \item ${(X \otimes Y)}^{T} = X^{T} \otimes Y^{T}$,
        \item $\text{vec}(XYZ) = (Z^{T}\otimes X)\text{vec}(Y)$,
    \end{enumerate}
    when matrix multiplications are well-defined.
\end{lemma}

By \cref{lemma:LSNormal,lemma:KronProp}, the normal equation of the LS
problem \cref{eq:LLLSDONFinal} is given as
\begin{equation*}\label{eq:NormalDON}
    \left[\left(\sum_{k=1}^{K} \epsilon_{k} T_k^{T} T_k\right) \otimes (B^{T}B)\right]
    \text{vec}(C^{T}) + \lambda \text{vec}(C^{T})
    = \sum_{k=1}^{K} \epsilon_{k} \left(T_k^{T} \otimes B^{T}\right)
    \text{vec}({F_k}),
\end{equation*}
or equivalently in matrix form
\begin{equation}\label{eq:NormalMatDON}
    B^{T} B C^{T} \left(\sum_{k=1}^{K} \epsilon_{k} T_k^{T}
    T_k\right) + \lambda C^{T}
    = B^{T} \left(\sum_{k=1}^{K} \epsilon_{k} F_k T_k\right).
\end{equation}

The matrix equation \cref{eq:NormalMatDON} is a special case of the generalized Sylvester
equation of type $A_1 X B_1 + A_2 X B_2 = C$~\cite{Chu1987}, where the algorithm for the
solution is a modification of the Bartels-Stewart algorithm~\cite{Bartels1972} for the Sylvester
equation $AX+XB=C$. Since $B^{T} B$ and $\sum_{k=1}^{K} \epsilon_{k} T_k^{T} T_k$ are
symmetric positive semi-definite matrices, they have spectral decompositions with nonnegative
eigenvalues. This makes it easy to solve the matrix equation \cref{eq:NormalMatDON}.

\begin{proposition}\label{prop:SPSDSyl}
    Let $A\in\mathbb{R}^{R\times R}$ and $B\in\mathbb{R}^{S\times S}$ be symmetric positive
    semi-definite matrices,
    $E\in\mathbb{R}^{R\times S}$ be any matrix, and $\lambda$ be a nonnegative real number.
    Then, the solution of the matrix equation
    \begin{equation}\label{eq:SylEq}
        AXB+\lambda X = E
    \end{equation}
    is given as
    \begin{equation}\label{eq:SylSol}
        X = Q_A \left[{(\mathbf{d}_A\mathbf{d}_B^{T}+\lambda \mathbf{1}_{R\times S})
        }^{\odot -1} \odot (Q_A^{T} E Q_B)\right] Q_B^{T},
    \end{equation}
    where $A = Q_A D_A Q_A^{T}$ and $B = Q_B D_B Q_B^{T}$ are the spectral
    decompositions with orthogonal matrices $Q_A$, $Q_B$ and diagonal
    matrices $D_A = \diag(\mathbf{d}_A)$, $D_B = \diag(\mathbf{d}_B)$
    when $\mathbf{d}_A\in\mathbb{R}^{R}$, $\mathbf{d}_B\in\mathbb{R}^{S}$. Here
    ${}^{\odot -1}$ denotes entrywise inverse
    and $\mathbf{1}_{R\times S}$ denotes the $R\times S$ matrix with every entry equal to one.
\end{proposition}
\begin{proof}
    For \cref{eq:SylEq}, multiplying $Q_A^{T}$ on the left and $Q_B$ on the right, we obtain
    \begin{equation}\label{eq:SylEqDiag}
        D_A Y D_B+\lambda Y = \tilde{E},
    \end{equation}
    where $Y = Q_A^{T} X Q_B$ and $\tilde{E} = Q_A^{T} E Q_B$.
    Comparing the $(r,s)$ components of \cref{eq:SylEqDiag}, we obtain $RS$ equations, each with
    one variable $Y_{r s}$:
    \begin{equation}\label{eq:SylSolo}
        d_{A,r} d_{B,s} Y_{r s} + \lambda Y_{r s} = {\tilde{E}}_{r s},
    \end{equation}
    where $d_{A,r}$ is the $r$-th entry of $\mathbf{d}_A$ and
    $d_{B,s}$ is the $s$-th entry of $\mathbf{d}_B$.
        We can write \cref{eq:SylSolo} in matrix form:
    \begin{equation}\label{eq:SylComp}
        {(\mathbf{d}_A\mathbf{d}_B^{T}+\lambda \mathbf{1}_{R\times S})} \odot Y
        = \tilde{E}.
    \end{equation}
    Since $\mathbf{d}_A\mathbf{d}_B^{T}+\lambda \mathbf{1}_{R\times S}$ is a matrix with
    positive entries, entrywise division is well-defined and rearranging \cref{eq:SylComp}
    yields \cref{eq:SylSol}.
\end{proof}

By \cref{prop:SPSDSyl}, we can find the last layer parameter
$C\in\mathbb{R}^{I\times J}$ that minimizes \cref{eq:DONFact} in matrix form:
\begin{equation*}\label{eq:CSolDON}
    C = {Q_T} \left[{(\mathbf{d}_T\mathbf{d}_B^{T}+\lambda \mathbf{1}_{I\times J})
    }^{\odot -1} \odot \left(Q_T^{T}
    \left(\sum_{k=1}^{K} \epsilon_{k} T_k^{T} F_k^{T} \right) B Q_B
    \right)\right]Q_B^{T},
\end{equation*}
where $B^{T} B = Q_B D_B Q_B^{T}$ and
$\sum_{k=1}^{K} \epsilon_{k} {T_k}^{T} T_k = Q_T D_T Q_T^{T}$ are the spectral
decompositions with orthogonal matrices $Q_B$, $Q_T$
and diagonal matrices $D_B = \diag(\mathbf{d}_B)$ and $D_T = \diag(\mathbf{d}_T)$.
This concludes the LS step for DeepONet. The LSGD method for DeepONet is described in
\cref{alg:LSGDDON}. Here, the LS step uses the full data batch, while the GD step
can utilize mini-batches.

\begin{algorithm}[tb]
    \caption{Hybrid Least Squares/Gradient Descent for DeepONet}\label{alg:LSGDDON}
    \begin{algorithmic}[1]
        \State{$(\theta^{B},\theta^{T})\gets (\theta_{0}^{B},\theta_{0}^{T})$:
        Initial parameters for the branch and the trunk}
        \State{$\theta^{L} \gets LS(\theta^{B},\theta^{T})$}
        \Comment{Solve the LS problem for $\theta^{L}$ over the full data batch}
        \For{$i=1,\dots$}
            \State{$(\theta^{B},\theta^{T}) \gets
            GD(\theta^{B},\theta^{T},\theta^{L})$}
            \Comment{Use a GD type optimizer to find
            $\theta^{B}$ and $\theta^{T}$ }
            \State{$\theta^{L} \gets LS(\theta^{B},\theta^{T})$}
            \EndFor{}
    \end{algorithmic}
\end{algorithm}

Note that if either $B^{T} B$ or $\sum_{k=1}^{K} \epsilon_{k} T_k^{T} T_k$ is singular or
ill-conditioned, the LS system \cref{eq:LexOrder} with $\lambda = 0$ either becomes
underdetermined and admits multiple minimizers, or numerical instability arises from division
of near-zero eigenvalues when computing $C$.
To avoid such a phenomenon, we set $\lambda>0$ to add Tikhonov
regularization~\cite{Golub1999,Tikhonov1963}, ensuring non-singularity for the LS system.
On the other hand, Cyr et al.~\cite{Cyr2020} suggested Box initialization for fully connected
neural networks with ReLU activation to reduce the likelihood of rank deficiency in
$B$ and $T_k$.

\section{Experimental results}\label{sec4}

In this section, we present experiments on various PDE problems to evaluate the proposed
hybrid training schemes, LS+Adam for DeepONets.
We report both supervised and unsupervised learning results, highlighting convergence speed
and solution accuracy.

We define LS+Adam as a practical LSGD method for DeepONets as follows.
Initially, we train all parameters using Adam for a moderate number of epochs.
This ensures that training with LS+Adam does not begin with poor initialization,
preventing it from reaching a bad local minimum in the LS step.
Then, we switch to the hybrid stage, where we use the LS step to optimize the last layer
parameters. After that, the LS step is applied once every few Adam epochs for the
hidden layer parameters; see \cref{alg:LSAdam} for LS+Adam.
These are similar to running Adam long enough in advance when using L-BFGS optimization,
typically in a PINN training.
However, using L-BFGS on DeepONet incurs significant memory overhead
because architectures are complex due to multiple neural networks, and the full-batch data
for each branch-trunk tuple must be used. This makes employing L-BFGS difficult
unless each network and data are small enough.

\begin{algorithm}[tb]
    \caption{LS+Adam for DeepONet}\label{alg:LSAdam}
    \begin{algorithmic}[1]
        \State{$(\theta^{B},\theta^{T},\theta^{L})\gets
        (\theta_{0}^{B},\theta_{0}^{T},\theta_{0}^{L})$: Initial parameters}

        \For{$i=1,\dots,I_0$}
            \State{$(\theta^{B},\theta^{T},\theta^{L}) \gets
            Adam(\theta^{B},\theta^{T},\theta^{L})$}
            \Comment{Initial Adam stage for all parameters}
        \EndFor{}

        \State{$\theta^{L} \gets LS(\theta^{B},\theta^{T})$}
        \Comment{Solve the LS problem for $\theta^{L}$ over the full data batch}

        \For{$i=1,\dots$} \Comment{Work unit block}
            \For{$j=1,\dots,J_0$}
                \State{$(\theta^{B},\theta^{T}) \gets
                Adam(\theta^{B},\theta^{T},\theta^{L})$}
                \Comment{Use Adam for hidden layer parameters}
            \EndFor{}
            \State{$\theta^{L} \gets LS(\theta^{B},\theta^{T})$}
        \EndFor{}
    \end{algorithmic}
\end{algorithm}

In each experiment, we use the Adam optimizer with a learning rate of $10^{-3}$ and
the first and second momentums $(\beta_1,\beta_2) = (0.99,0.999)$.
We also allow the Adam momentums before the LS step to be maintained after the LS step
to ensure stable training.
He normal initialization~\cite{He2015} is used for parameter initialization.
We adopt the Swish function $x/(1+e^{-x})$ as the activation function in all
experiments because we found that it yielded better results than other activation
functions such as ReLU and tanh. For further information, refer
to~\cite{Lu2024,Ramachandran2017}.
For training with Adam-only, no regularization term for the last layer parameters is used,
but for training with LS+Adam, this regularization term is used with a positive weight
$\lambda$.
In general, a small $\lambda$ is used when the model has complex structures or the loss
function contains a PI-loss term for unsupervised learning.

In \cref{subsec41,subsec42,subsubsec441,subsubsec442},
we address DeepONets with supervised learning.
Additionally, \cref{subsec41,subsubsec442,subsubsec443}
cover unsupervised learning for advection equation with constant
coefficient and the 2D Poisson equation where the input function is either a source term or
boundary condition~(BC)\@.

For DeepONet training, the training and validation data consist of 1,000 and 100 functions,
respectively, and Adam step uses a batch consisting of 50 functions.
For the LS+Adam hybrid stage,
we define one Work Unit~(WU) as a cycle of five Adam epochs followed by one LS step in
\cref{alg:LSAdam}.
We empirically found that using five Adam epochs per LS step balances
computational cost and convergence behavior well in most experiments.
Before entering the LS+Adam stage, we train all parameters using Adam for 500 epochs,
equivalent to 100 WUs, followed by an LS step.
Experimental results show that the training time for a single WU in the hybrid stage is
2--5\% longer than that of Adam's.

For the loss in supervised learning, we use
\begin{equation*}\label{eq:Loss_Sup}
    \epsilon L_{\text{data}} +
    \lambda {\|C\|}_F^2,
\end{equation*}
where ${\| \cdot \|}_{F}$ denotes the Frobenius norm, $\epsilon=1$ and
$L_{\text{data}}$ is the $L^2$ mean squared error~(MSE) on the given data pairs
$(\hat{u}_d,\hat{y}_d)$.
For the loss in unsupervised learning, we define a PI-loss:
\begin{equation*}\label{eq:Loss_Unsup}
    \epsilon_{1} L_{\text{data}} +
    \epsilon_{2} L_{\text{physics}} +
    \lambda {\|C\|}_F^2,
\end{equation*}
where $\epsilon_{1} = 1$,
$L_{\text{data}}$ is the $L^2$ MSE on the data pairs $(\hat{u}_{d_1},\hat{y}_{d_1})$
where $\hat{y}_{d_1}$ corresponds to the initial or boundary conditions of the governing PDE,
and $L_{\text{physics}}$ is the $L^2$ MSE of the PDE residuals
$(\hat{u}_{d_2},\hat{y}_{d_2})$ where $\hat{y}_{d_2}$ corresponds to
the interior points to compute residual.

Note that we cannot apply our method to unsupervised learning in
\cref{subsec42,subsubsec441} because the operator used in the physics term is nonlinear in
\cref{subsec42}, and in \cref{subsubsec441}, although it is linear,
it does not satisfy the condition \cref{eq:CondOpDON} because it depends on the input
function.

Refer to \cref{Tab:DON_config_Sup,Tab:DON_config_Unsup}
and corresponding sections for each model structure and
training details. All computations were performed using Google JAX~\cite{jax2018} on a machine
with Intel Xeon Gold 6430 processors and NVIDIA GeForce RTX 4090 with 24 GB memory.

\begin{table}
    \caption{\textit{Input functions, network structures, and regularization weights
    of DeepONet models with supervised learning.}
    IC and BC stand for initial condition and boundary condition.
    FCN and CNN stand for fully connected network and convolutional
    neural network, respectively. The CNN structure is described in \cref{subsubsec441}.
    Swish activation is used on all branches and trunks.}
    \centering
    {\footnotesize
    \begin{tabular}{cclllc}
        \toprule
        \multirow{2}{*}{Equation} & \multirow{2}{*}{Section} &
        {Input} &
        \multirow{2}{*}{\begin{tabular}{l}Branch\\structure\end{tabular}} &
        \multirow{2}{*}{\begin{tabular}{l}Trunk\\structure\end{tabular}} &
        \multirow{2}{*}{$\lambda$}\\
        & & {function} & & & \\\midrule
        \multirow{2}{*}{Advection} &
        \multirow{2}{*} {\nolinebreak\ref{subsec41}} &
        \multirow{2}{*}{BC+IC} &
        \multirow{2}{*}{\begin{tabular}{l}FCN $[65$,\\$100, 100, 100]$\end{tabular}} &
        \multirow{2}{*}{\begin{tabular}{l}FCN $[2$,\\$100, 100, 100]$\end{tabular}} &
        \multirow{2}{*}{$10^{-6}$}\\\\
        \midrule
        \multirow{2}{*}{\begin{tabular}{c}Diffusion-\\Reaction\end{tabular}} &
        \multirow{2}{*}{\nolinebreak\ref{subsec42}} &
        \multirow{2}{*}{Source} &
        \multirow{2}{*}{\begin{tabular}{l}FCN $[33$,\\$100, 100, 100]$\end{tabular}} &
        \multirow{2}{*}{\begin{tabular}{l}FCN $[2$,\\$100, 100, 100]$\end{tabular}} &
        \multirow{2}{*}{$10^{-6}$}\\\\\midrule
        \multirow{4.4}{*}{Poisson}& \multirow{2}{*}{\nolinebreak\ref{subsubsec441}} &
        \multirow{2}{*}{Coefficient} &
        \multirow{2}{*}{\begin{tabular}{l}CNN + FCN\\$[1024, 150, 150]$\end{tabular}} &
        \multirow{2}{*}{\begin{tabular}{l}FCN $[2$,\\$150, 150, 150]$\end{tabular}} &
        \multirow{2}{*}{$10^{-9}$}\\\\\cmidrule(lr){2-6}
        & \multirow{2}{*}{\nolinebreak\ref{subsubsec442}} &
        \multirow{2}{*}{BC} &
        \multirow{2}{*}{\begin{tabular}{l}FCN $[129$,\\$150, 150, 150]$\end{tabular}} &
        \multirow{2}{*}{\begin{tabular}{l}FCN $[2$,\\$150, 150, 150]$\end{tabular}} &
        \multirow{2}{*}{$10^{-6}$}\\\\\bottomrule
    \end{tabular}~\label{Tab:DON_config_Sup}}
\end{table}

\begin{table}
    \caption{\textit{Input functions, network structures, regularization weights,
    and physics term weights of DeepONet models with unsupervised learning.}
    The arrow indicates a gradual exponential decrease of the last layer parameter
    regularization weight, starting at 100 WU up to 1,000 WU\@.
    The CNN structure is described in \cref{subsubsec443}.
    Swish activation is used on all branches and trunks.
    }
    \centering
    {\footnotesize
    \begin{tabular}{cclllcc}
        \toprule
        \multirow{2}{*}{Equation} & \multirow{2}{*}{Section} &
        {Input} &
        \multirow{2}{*}{\begin{tabular}{l}Branch\\structure\end{tabular}} &
        \multirow{2}{*}{\begin{tabular}{l}Trunk\\structure\end{tabular}} &
        \multirow{2}{*}{$\lambda$} & \multirow{2}{*}{$\epsilon_2$}\\
        & & {function} & & & \\\midrule
        \multirow{2}{*}{Advection} &
        \multirow{2}{*} {\nolinebreak\ref{subsec41}} &
        \multirow{2}{*}{BC+IC} &
        \multirow{2}{*}{\begin{tabular}{l}FCN $[65$,\\$100, 100, 100]$\end{tabular}} &
        \multirow{2}{*}{\begin{tabular}{l}FCN $[2$,\\$100, 100, 100]$\end{tabular}} &
        \multirow{2}{*}{$10^{-6}$} & \multirow{2}{*}{$10^{-1}$}\\\\\midrule
        \multirow{4.4}{*}{Poisson} & \multirow{2}{*}{\nolinebreak\ref{subsubsec442}} &
        \multirow{2}{*}{BC} &
        \multirow{2}{*}{\begin{tabular}{l}FCN $[129$,\\$150, 150, 150]$\end{tabular}} &
        \multirow{2}{*}{\begin{tabular}{l}FCN $[2$,\\$150, 150, 150]$\end{tabular}} &
        \multirow{2}{*}{$10^{-9}$} & \multirow{2}{*}{$10^{-4}$}\\\\\cmidrule(lr){2-7}
        & \multirow{2}{*}{\nolinebreak\ref{subsubsec443}} &
        \multirow{2}{*}{Source} &
        \multirow{2}{*}{\begin{tabular}{l}CNN + FCN\\$[1024,150,150]$\end{tabular}} &
        \multirow{2}{*}{\begin{tabular}{l}FCN $[2$,\\$150, 150, 150]$\end{tabular}} &
        {$10^{-9}\to$} & \multirow{2}{*}{$10^{-4}$}\\
        & & & & & {$10^{-14}$}&\\\bottomrule
    \end{tabular}~\label{Tab:DON_config_Unsup}}
\end{table}

\begin{figure}
  \centering
  \begin{subfigure}{.33\textwidth}
      \centering
      \includegraphics[width=\linewidth]{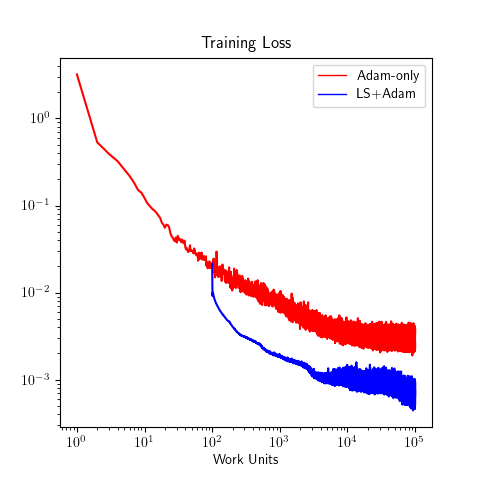}\label{fig:Advec1_Loss}
  \end{subfigure}%
  \begin{subfigure}{.33\textwidth}
      \centering
      \includegraphics[width=\linewidth]{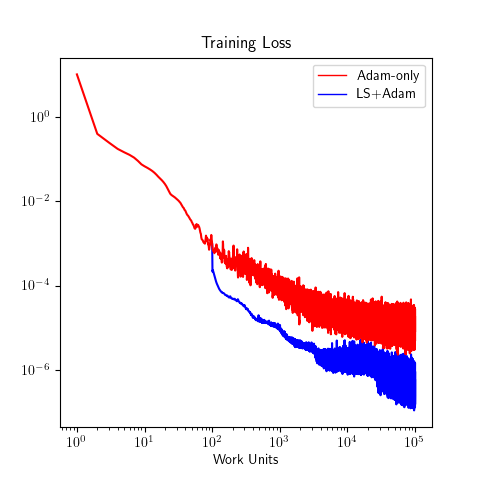}\label{fig:Advec1U_Loss}
  \end{subfigure}%
  \begin{subfigure}{.33\textwidth}
      \centering
      \includegraphics[width=\linewidth]{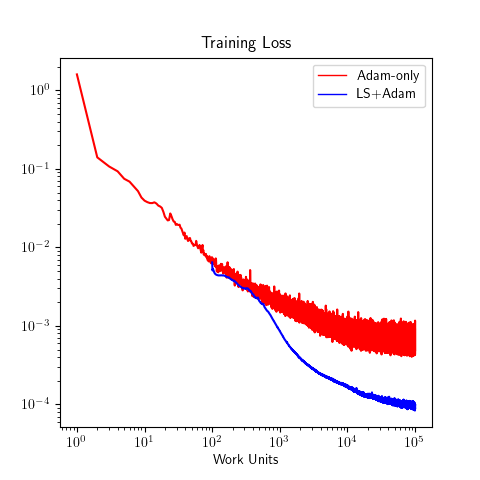}\label{fig:DiffReact_Loss}
  \end{subfigure}
  \begin{subfigure}{.33\textwidth}
      \centering
      \includegraphics[width=\linewidth]{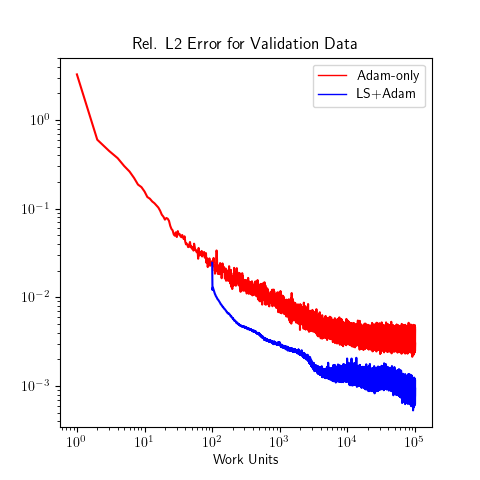}
      \caption{Constant-coefficient\\advection, supervised}\label{fig:Advec1_RelL2}
  \end{subfigure}%
  \begin{subfigure}{.33\textwidth}
      \centering
      \includegraphics[width=\linewidth]{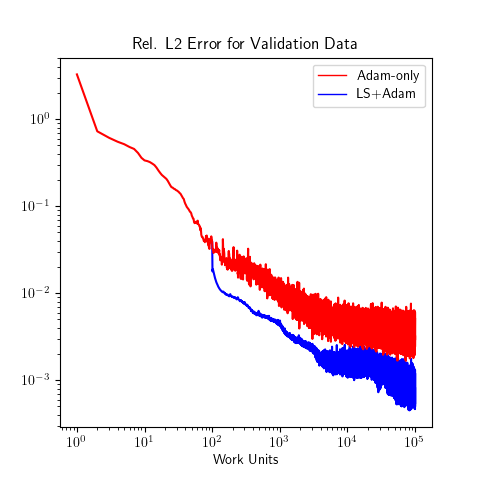}
      \caption{Constant-coefficient\\advection, unsupervised}\label{fig:Advec1U_RelL2}
  \end{subfigure}%
  \begin{subfigure}{.33\textwidth}
      \centering
      \includegraphics[width=\linewidth]{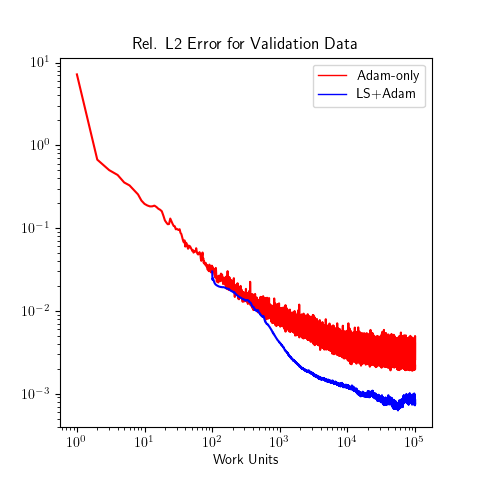}
      \caption{Diffusion-reaction\\with source, supervised}\label{fig:DiffReact_RelL2}
  \end{subfigure}
  \caption{\textit{Solving advection and diffusion-reaction equations
  via DeepONet: Adam-only vs. LS+Adam.}
  Top and bottom rows denote the loss $L_{\text{data}}$ for (a) and (c), and
  $ L_{\text{data}}+ \epsilon_2 L_{\text{physics}}$ for (b) in training and the
  mean relative $L^2$ error for 100 validation data, respectively.
  Red and blue graphs represent the cases of Adam-only and LS+Adam, respectively.
  }\label{fig:SupLoss1}
\end{figure}

\subsection{Advection equation with constant coefficient}\label{subsec41}
Consider a 1D advection equation:
\begin{equation*}\label{eq:Advec}
\begin{alignedat}{3}
    \frac{\partial u}{\partial t} + a(x)\frac{\partial u}{\partial x} &= 0,\quad
    &&(x,t)\in{(0,1]}^2,\\ 
    u(x,0) &= P(x),\quad &&x\in[0,1],\\ 
    u(0,t) &= Q(t),\quad &&t\in[0,1], 
\end{alignedat}
\end{equation*}
where $a\in L^{\infty}([0,1];{\mathbb{R}}_{>0})$ and $P(0)=Q(0)$.
We aim to learn a solution operator via DeepONet, which maps the initial data $P$ and
the boundary data $Q$ to $u$, where the coefficient is given as a constant function
$a(x)\equiv a$. Both supervised and unsupervised learning are used.

Note that the analytical solution is
\begin{equation*}\label{eq:ConstAdvecSol}
    u^{*}(x,t) = \begin{cases}
    P(x-at), \quad &x-at \geq 0,\\
    Q(t-\frac{x}{a}), \quad &x-at < 0.
\end{cases}
\end{equation*}
Since the solution may introduce a non-differentiable cusp along the line $x-at=0$,
it is challenging to generate such a solution by automatic differentiation for the PI-loss
in unsupervised learning. To avoid this difficulty, we introduce an additional condition
$P'(0) = -\frac{1}{a}Q'(0)$, which restricts the solution to be differentiable along the
line $x-at=0$.
Furthermore, instead of having two separate input functions, $P$ and $Q$, we can naturally
concatenate $P$ and $Q$ into one input function along the domains of IC and BC\@.
Let $\mathbf{p}=[P(0)\cdots P(1)]$ and $\mathbf{q}=[Q(0)\cdots Q(1)]$ be the discretizations of
$P$ and $Q$ along their domains, $[0,1]\times\{0\}$ and $\{0\}\times[0,1]$, respectively.
Since $P(0)=Q(0)$, we can concatenate $\mathbf{p}$ and $\mathbf{q}$ by
$\mathbf{r}=[Q(1)\cdots Q(0)=P(0) \cdots P(1)]$, where $\mathbf{q}$ is flipped and one of the
duplicate values $P(0)$ or $Q(0)$ is removed.

To generate such input functions $P$ and $Q$, we first sample $f$ from a Gaussian process~(GP)
in the interval $[-a,1]$ with zero mean and a squared exponential covariance kernel
\begin{equation}\label{eq:SqExpKer}
    k(x_1,x_2) = \sigma^2 \text{exp}\left(-\frac{{|x_1 - x_2|}^2}{2l^2}\right)
\end{equation}
having scale parameter $l=0.2$ and variance $\sigma^2 = 1$.
Then, we set
\begin{equation*}\label{eq:AdvecInputs}
    \begin{alignedat}{3}
        P(x) &= f(x),\quad &&x \in [0,1],\\
        Q(t) &= f(-at),\quad &&t \in [0,1].
    \end{alignedat}
\end{equation*}
The generated $P$ and $Q$ satisfy $P(0)=Q(0)$ and $P'(0) = -\frac{1}{a}Q'(0)$.
In this problem, we choose $a = 0.5$.
The original input functions $P$ and $Q$ are discretized at $33$ equidistant grid points
of $[0,1]$, and the concatenated input $\mathbf{r}$ is a vector of length $65$.
The output function is evaluated on $33\times33$ equidistant grid points of ${[0,1]}^2$.

As shown in \cref{fig:SupLoss1}\nolinebreak(a) and (b), training with LS+Adam reduces the
training loss faster and significantly improves the model's performance compared to training
with Adam-only in both supervised and unsupervised learning cases.
In both cases, the mean relative $L^2$ error of LS+Adam at 10,000 WU is much smaller than that
of Adam-only at 100,000 WU\@.

\begin{figure}
    \centering
    \begin{subfigure}{.4\textwidth}
        \centering
        \includegraphics[width=\linewidth]{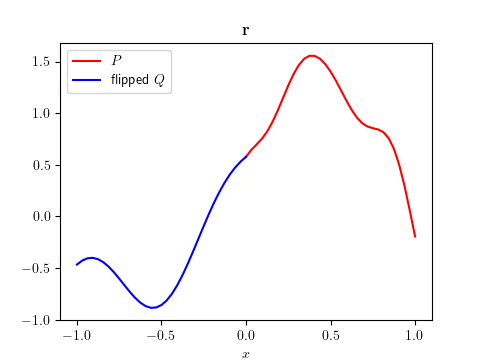}
        \caption{}\label{fig:Advec1_Input}
    \end{subfigure}%
    \begin{subfigure}{.4\textwidth}
        \centering
        \includegraphics[width=\linewidth]{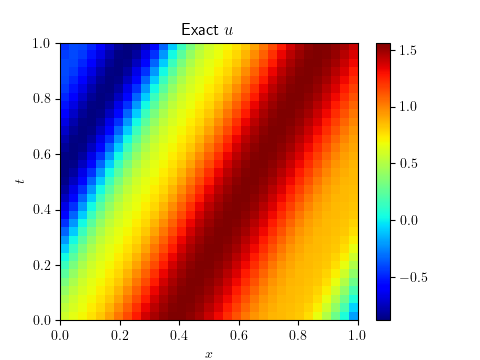}
        \caption{}\label{fig:Advec1_Exact}
    \end{subfigure}
    \begin{subfigure}{.4\textwidth}
        \centering
        \includegraphics[width=\linewidth]{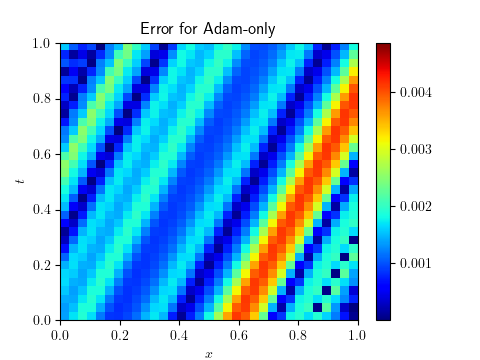}
        \includegraphics[width=\linewidth]{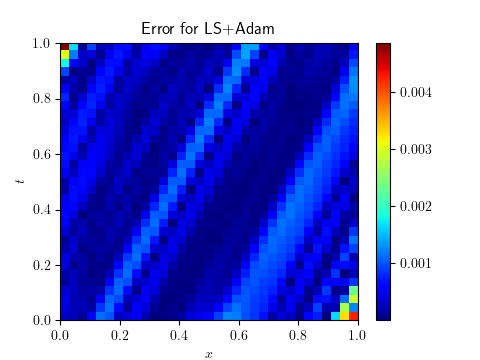}
        \caption{Supervised }\label{fig:Advec1_ErrorAdam_3}
    \end{subfigure}%
    \begin{subfigure}{.4\textwidth}
        \centering
        \includegraphics[width=\linewidth]{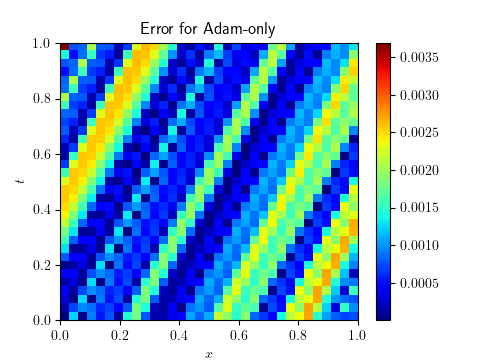}
        \includegraphics[width=\linewidth]{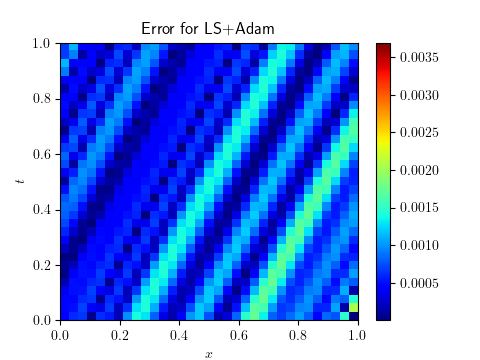}
        \caption{Unsupervised }\label{fig:Advec1_ErrorLSAdam_2}
    \end{subfigure}
    \caption{\textit{Advection equation with constant coefficient: Adam-only vs.
    LS+Adam, supervised and unsupervised.}
    (a)~Test data: concatenated input function $\mathbf{r}$ of IC and BC\@.
    (b)~Exact solution.
    (c), (d)~Top: absolute error of the DeepONet solution with Adam-only at $10^5$ WU\@.
    Bottom: absolute error of the DeepONet solution with LS+Adam at $10^4$ WU\@.
    }\label{fig:Advec1_TestData}
\end{figure}

\Cref{fig:Advec1_TestData} illustrates the solution errors of DeepONet trained with
Adam-only and LS+Adam for unseen test data in supervised and unsupervised learning.
In the supervised learning, the $L^2$ error of Adam-only at 100,000 WU is $\num{1.83e-3}$,
while the $L^2$ error of LS+Adam at 10,000 WU is $\num{5.94e-4}$.
In the unsupervised learning, the $L^2$ error of Adam-only at 100,000 WU is $\num{1.30e-3}$,
whereas the $L^2$ error of LS+Adam at 10,000 WU is $\num{7.77e-4}$.
All cases show that the errors tend to align along the lines parallel to $x-0.5t=0$.
Additionally, the solutions from LS+Adam are likely to exhibit relatively large errors at the
corners $(0,1),(1,0)$.

\subsection{Diffusion-reaction equation with a source term}\label{subsec42}
We consider a diffusion-reaction equation with a source
term $f(x)$ and zero initial and boundary conditions:
\begin{equation*}\label{eq:DiffReact}
    \begin{alignedat}{3}
        \frac{\partial u}{\partial t} = \frac{\partial}{\partial x} \left(D(x)
        \frac{\partial u}{\partial x}\right) &+ ku^2 + f(x),\quad
        &&(x,t)\in(0,1)\times(0,1],\\ 
        u(x,0) &= 0,\quad &&x\in(0,1),\\ 
        u(0,t) = u(1,t) &= 0,\quad &&t\in(0,1], 
    \end{alignedat}
\end{equation*}
where $D(x) \equiv 0.01$ and $k = 1$.
We aim to learn a solution operator via DeepONet,
which maps the source $f$ to the solution $u$. The input source $f$ is generated from
a GP with zero mean and a squared exponential covariance kernel \cref{eq:SqExpKer},
whose parameters are given by $l=0.2$ and $\sigma^2 = 0.5$.
The input function $f$ is discretized at $33$ equidistant grid points of $[0,1]$ and
the output function is evaluated on $33\times33$ equidistant grid points of
${[0,1]}^2$.
The reference solutions are generated by the finite
difference method~(FDM) that is implicit in time and central in space~\cite{Wang2021}.

Similar to the experiments in \cref{subsec41}, \cref{fig:SupLoss1}\nolinebreak(c) shows that
LS+Adam outperforms Adam-only in both training loss decay and model performance.
Looking at the mean relative $L^2$ error, LS+Adam gives a much smaller error level
at 10,000 WU than the error level at 100,000 WU when Adam-only is applied.

\begin{figure}
    \centering
    \begin{subfigure}{.4\textwidth}
        \centering
        \includegraphics[width=\linewidth]{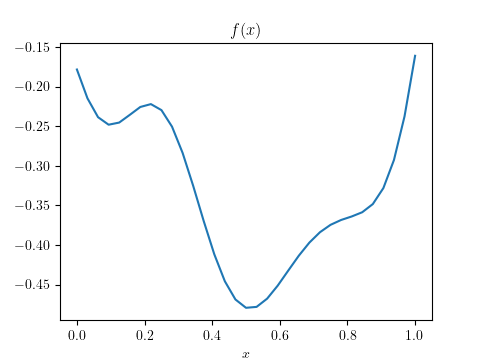}
        \caption{}\label{fig:DiffReact_Input}
    \end{subfigure}%
    \begin{subfigure}{.4\textwidth}
        \centering
        \includegraphics[width=\linewidth]{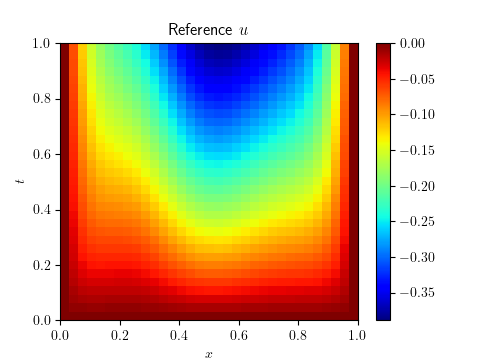}
        \caption{}\label{fig:DiffReact_Exact}
    \end{subfigure}
    \begin{subfigure}{.4\textwidth}
        \centering
        \includegraphics[width=\linewidth]{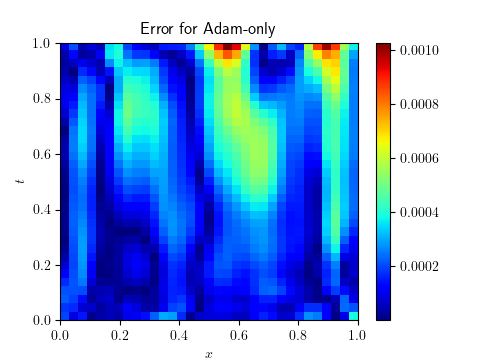}
        \caption{}\label{fig:DiffReact_ErrorAdam_3}
    \end{subfigure}%
    \begin{subfigure}{.4\textwidth}
        \centering
        \includegraphics[width=\linewidth]{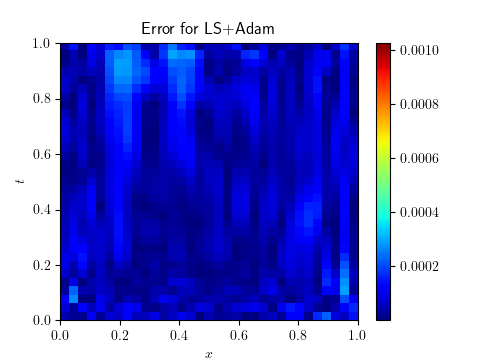}
        \caption{}\label{fig:DiffReact_ErrorLSAdam_2}
    \end{subfigure}
    \caption{\textit{Diffusion-reaction equation with a source term: Adam-only vs.
    LS+Adam.}
    (a)~Test data: input initial condition $f(x)$.
    (b)~Reference solution.
    (c)~Absolute error of the DeepONet solution with Adam-only at $10^5$ WU\@.
    (d)~Absolute error of the DeepONet solution with LS+Adam at $10^4$ WU\@.
    }\label{fig:DiffReact_TestData}
\end{figure}

\Cref{fig:DiffReact_TestData} shows the solution errors of DeepONet trained with
Adam-only and LS+Adam for unseen test data.
The $L^2$ error of Adam-only at 100,000 WU is $\num{2.78e-4}$, while the $L^2$ error of
LS+Adam at 10,000 WU is $\num{8.48e-5}$.
In both cases, the error tends to increase as $t$ grows.

\subsection{2D Poisson equation}\label{subsec44}
In this section, we consider a 2D Poisson equation on the unit square with Dirichlet BC:\@
\begin{equation}\label{eq:Poisson}
    \begin{alignedat}{3}
        -\nabla\cdot\left(\kappa\nabla u\right) &= f ,\quad &&(x,y)\in\Omega={(0,1)}^2,\\
        u &= g,\quad &&(x,y)\in\partial\Omega.
    \end{alignedat}
\end{equation}
The solution operator via DeepONet takes as input either the coefficient $\kappa$, the
Dirichlet boundary condition $g$, or the source term $f$, depending on the model problem.
In \cref{subsubsec441,subsubsec442}, supervised learning will be performed.
On the other hand, in \cref{subsubsec442,subsubsec443}, unsupervised learning with
the previously mentioned PI-loss will be performed.
The reference solutions are generated by the finite difference method on finer grids of size
$129 \times 129$.

\begin{figure}
    \centering
    \begin{subfigure}{.33\textwidth}
        \centering
        \includegraphics[width=\linewidth]{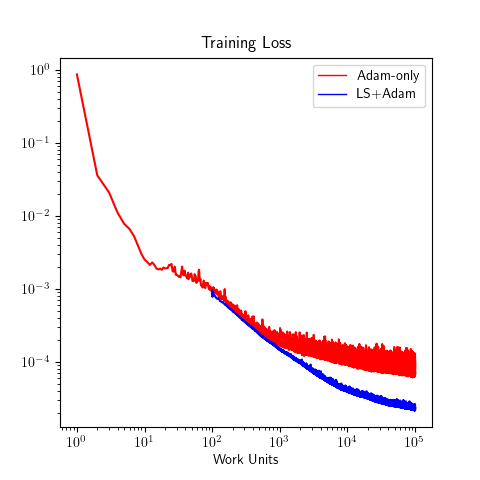}\label{fig:PoissonK_Loss}
    \end{subfigure}%
    \begin{subfigure}{.33\textwidth}
        \centering
        \includegraphics[width=\linewidth]{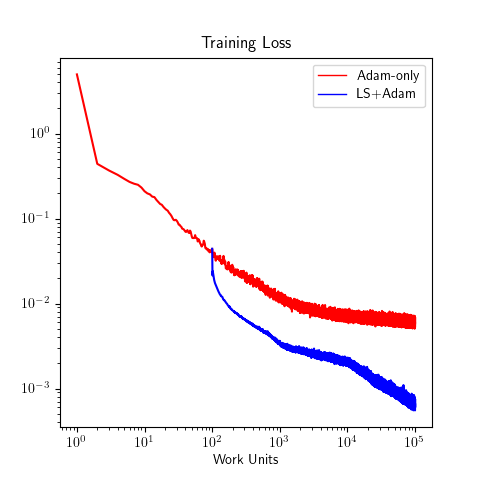}\label{fig:PoissonG_Loss}
    \end{subfigure}
    \begin{subfigure}{.33\textwidth}
        \centering
        \includegraphics[width=\linewidth]{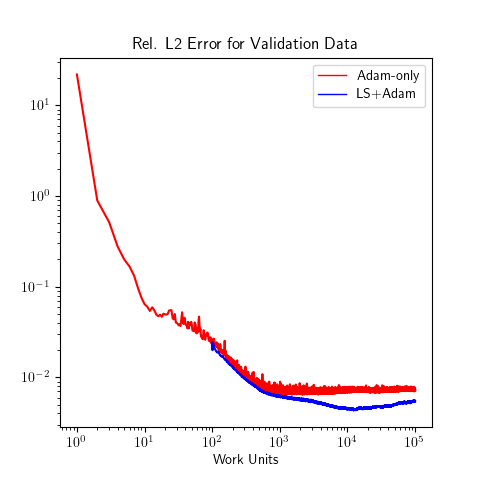}
        \caption{Variable coefficient }\label{fig:PoissonK_RelL2}
    \end{subfigure}%
    \begin{subfigure}{.33\textwidth}
        \centering
        \includegraphics[width=\linewidth]{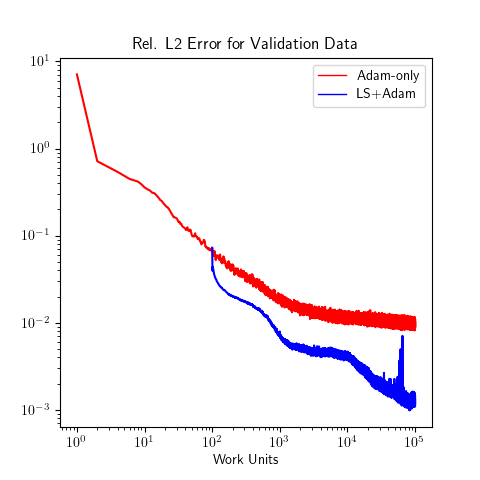}
        \caption{Variable BC }\label{fig:PoissonG_RelL2}
    \end{subfigure}
    \caption{\textit{Solving Poisson equation with supervised learning: Adam-only vs. LS+Adam}
    Top and bottom rows denote the loss $L_{\text{data}}$ in training and the mean
    relative $L^2$ error for 100 validation data, respectively.
    Red and blue graphs denote the cases of Adam-only and LS+Adam, respectively.
    }\label{fig:SupLoss3}
\end{figure}

\subsubsection{Poisson equation with a variable coefficient}\label{subsubsec441}
In this section, we learn a solution operator from a 2D coefficient $\kappa$ to $u$ via
DeepONet when $f=1$ and $g=0$.

The input function $\kappa$ is defined as
$\kappa = \exp(\tilde{\kappa})$, where $\tilde{\kappa}$ is generated from
a GP with zero mean and a 2D squared exponential covariance kernel
\begin{equation}\label{eq:SqExpKer2D}
    k(x_1,x_2,y_1,y_2) = \sigma^2 \text{exp}\left(-\frac{{|x_1 - x_2|}^2}{2{l_x}^2}
    -\frac{{|y_1 - y_2|}^2}{2{l_y}^2}\right),
\end{equation}
having scale factor $l_x = l_y = 0.1$ and variance $\sigma^2 = 0.2$.
The input function is a $32\times 32$ 2D image from the discretization at the center of
each square cell generated by an equidistant $33 \times 33$ grid of ${[0,1]}^2$,
and the output function is evaluated at $33\times33$ equidistant grid points of
${[0,1]}^2$.
The CNN of the branch network consists of three layers with $2\times 2$ kernels with
$2 \times 2$ strides. Since the channel sizes are $[1,16,32,64]$, the output of the CNN
is a $64$-channel $4\times 4$ image, which is then flattened and becomes the input of the FCN
part of the branch.

\Cref{fig:SupLoss3}\nolinebreak(a) shows that LS+Adam still
outperforms Adam-only in terms of training loss reduction and model performance when the
input function is given as a 2D image.
In this experiment, Adam-only exhibits early overfitting behavior where the relative $L^2$ error
is stagnant after about 1,000 WU, while Adam+LS shows overfitting after 10,000 WU\@.

\begin{figure}
    \centering
    \begin{subfigure}{.4\textwidth}
        \centering
        \includegraphics[width=\linewidth]{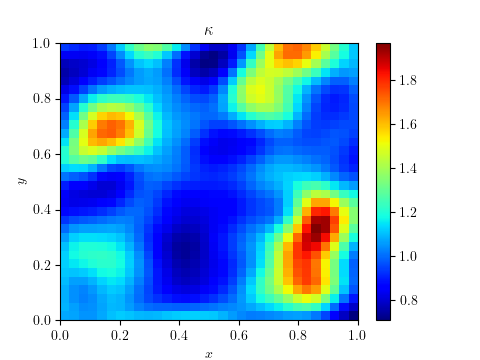}
        \caption{}\label{fig:PoissonK_Input}
    \end{subfigure}%
    \begin{subfigure}{.4\textwidth}
        \centering
        \includegraphics[width=\linewidth]{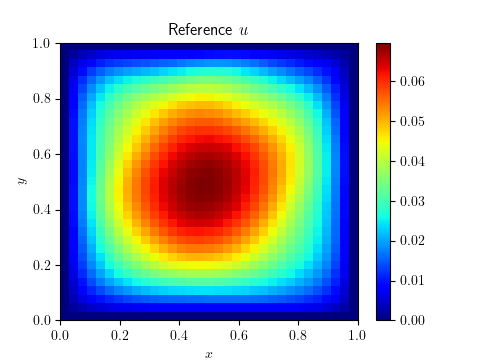}
        \caption{}\label{fig:PoissonK_Exact}
    \end{subfigure}
    \begin{subfigure}{.4\textwidth}
        \centering
        \includegraphics[width=\linewidth]{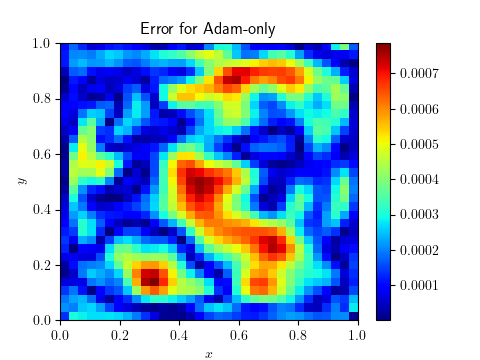}
        \caption{}\label{fig:PoissonK_ErrorAdam_3}
    \end{subfigure}%
    \begin{subfigure}{.4\textwidth}
        \centering
        \includegraphics[width=\linewidth]{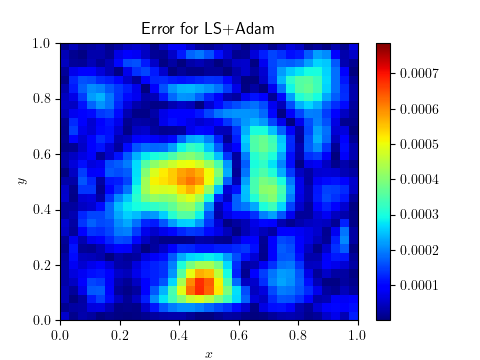}
        \caption{}\label{fig:PoissonK_ErrorLSAdam_2}
    \end{subfigure}
    \caption{\textit{Poisson equation with a variable coefficient: Adam-only vs.
    LS+Adam.}
    (a)~Test data: input coefficient $\kappa(x,y)$.
    (b)~Reference solution.
    (c)~Absolute error of the DeepONet solution with Adam-only at $10^5$ WU\@.
    (d)~Absolute error of the DeepONet solution with LS+Adam at $10^4$ WU\@.
    }\label{fig:PoissonK_TestData}
\end{figure}

\Cref{fig:PoissonK_TestData} shows the solution errors of DeepONet with Adam-only and
LS+Adam for a test data.
The $L^2$ error of Adam-only at 100,000 WU is $\num{3.23e-4}$, while the $L^2$ error of
LS+Adam at 10,000 WU is $\num{1.89e-4}$.

\begin{figure}
    \centering
    \begin{subfigure}{.33\textwidth}
        \centering
        \includegraphics[width=\linewidth]{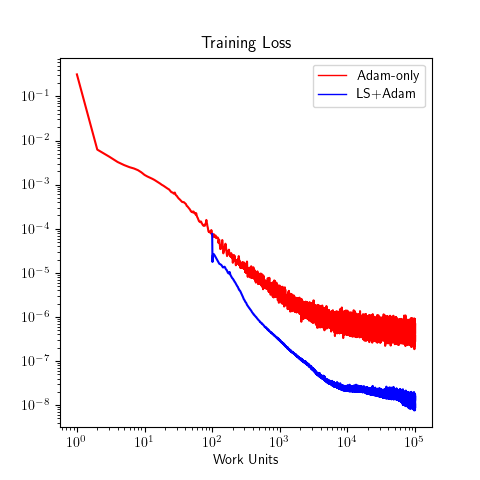}\label{fig:PoissonGU_Loss}
    \end{subfigure}%
    \begin{subfigure}{.33\textwidth}
        \centering
        \includegraphics[width=\linewidth]{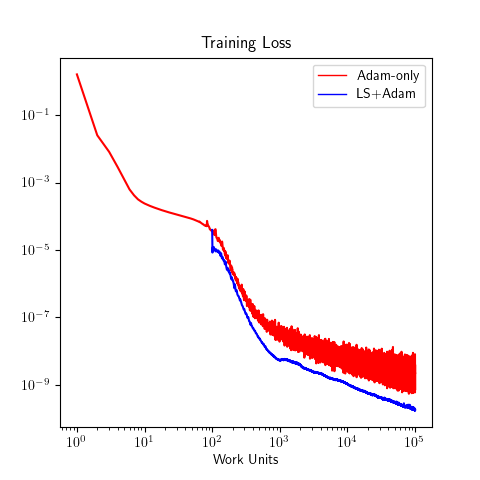}\label{fig:PoissonFU_Loss}
    \end{subfigure}
    \begin{subfigure}{.33\textwidth}
        \centering
        \includegraphics[width=\linewidth]{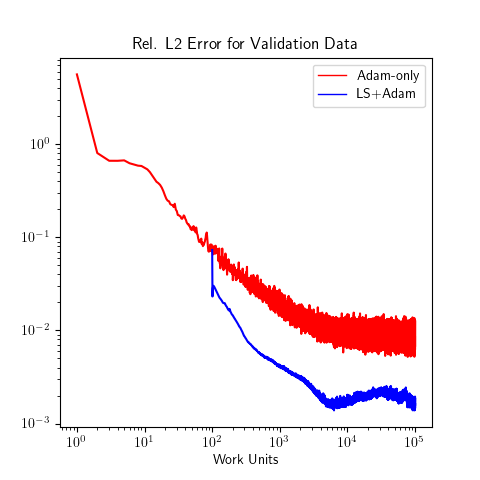}
        \caption{Variable BC }\label{fig:PoissonGU_RelL2}
    \end{subfigure}%
    \begin{subfigure}{.33\textwidth}
        \centering
        \includegraphics[width=\linewidth]{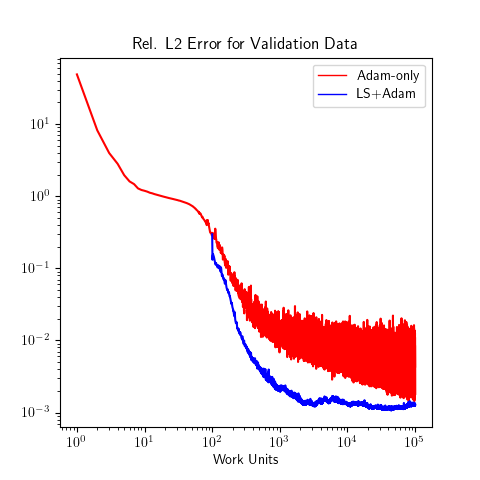}
        \caption{Variable source }\label{fig:PoissonFU_RelL2}
    \end{subfigure}
    \caption{\textit{Solving Poisson equation with unsupervised learning: Adam-only vs.
    LS+Adam}.
    Top and bottom rows denote the PI-loss
    $L_{\text{data}}+\epsilon_2 L_{\text{physics}}$
    in training and mean relative $L^2$ error for 100 validation data, respectively.
    Red and blue graphs denote the cases of Adam-only and LS+Adam, respectively.
    }\label{fig:UnSupLoss1}
\end{figure}

\subsubsection{Poisson equation with a variable boundary}\label{subsubsec442}
In this section, we learn a solution operator from the Dirichlet BC $g$ to $u$ via DeepONet
when $\kappa=1$ and $f=0$. The input function $\tilde{g}$ is a 1D function
defined on the interval $[0,4]$ generated from a GP with zero mean and a periodic
covariance kernel
\begin{equation*}\label{eq:PeriodicKer}
    k(x_1,x_2) = \sigma^2 \text{exp}\left(-\frac{2}{{l}^2}\sin^{2}
    {\left(\frac{\pi|x_1 - x_2|} {p}\right)}\right),
\end{equation*}
where the scale factor, the period, and the variance are $l = 0.3$, $p = 4$, and
$\sigma^2 = 1$, respectively,
so that $g(\mathbf{h}(t)) = \tilde{g}(t)$ where
$\mathbf{h}\colon[0,4]\to\partial\Omega$ is the arc length parametrization of $\partial\Omega$
such that
\begin{equation*}\label{eq:PoisFlatBC}
    \begin{alignedat}{3}
        \mathbf{h}(t) &=
        \begin{cases}
            (t,0),&t\in[0,1),\\ 
            (1,t-1),&t\in[1,2),\\ 
            (3-t,1),&t\in[2,3),\\ 
            (0,4-t),&t\in[3,4]. 
        \end{cases}
    \end{alignedat}
\end{equation*}
The input function $\tilde{g}$ is discretized by $129$ equidistant grid points of $[0,4]$,
and the output function is evaluated on $33\times33$ equidistant grid points of
${[0,1]}^2$.
For the unsupervised learning case, we found that using input $\tilde{g}$ with scale factor
$0.1$ is more effective than using the original input $\tilde{g}$ for both Adam-only and LS+Adam
training. In this case, the reference solution $u$ is also reduced to $0.1$ times the
original solution, due to the linearity of the PDE\@. Note that this is the same as
generating $\tilde{g}$ with $\sigma^2 = 0.1$.

\Cref{fig:SupLoss3}\nolinebreak(b) and~\cref{fig:UnSupLoss1}\nolinebreak(a)
show that in both supervised and unsupervised learning,
LS+Adam outperforms Adam-only in terms of training loss reduction and model performance.
Here, LS+Adam achieves the error level around 1,000 WU, which Adam achieves at 100,000 WU\@.

\begin{figure}
    \centering
    \begin{subfigure}{.4\textwidth}
        \centering
        \includegraphics[width=\linewidth]{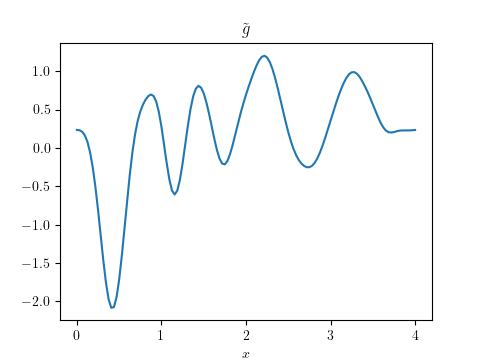}
        \caption{}\label{fig:PoissonG_Input}
    \end{subfigure}%
    \begin{subfigure}{.4\textwidth}
        \centering
        \includegraphics[width=\linewidth]{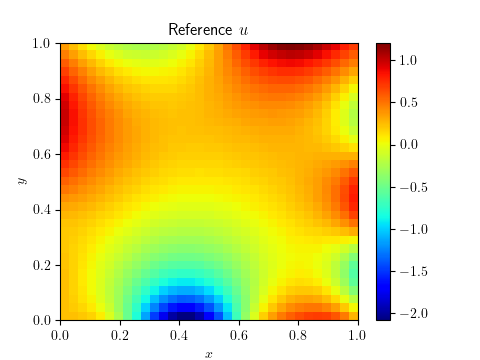}
        \caption{}\label{fig:PoissonG_Exact}
    \end{subfigure}
    \begin{subfigure}{.4\textwidth}
        \centering
        \includegraphics[width=\linewidth]{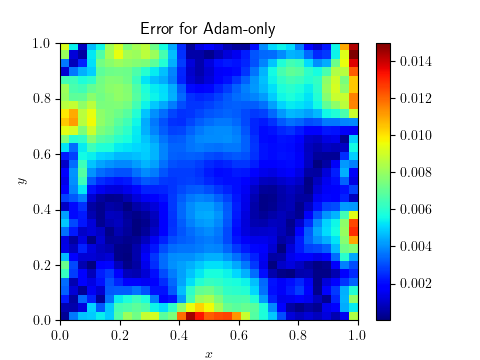}
        \includegraphics[width=\linewidth]{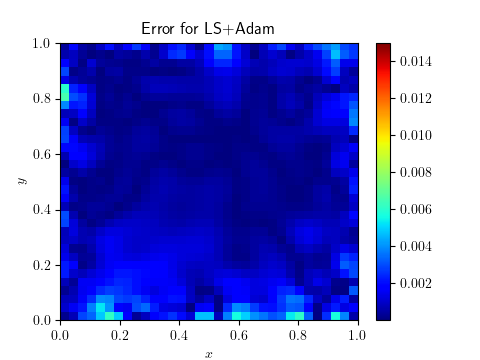}
        \caption{Supervised }\label{fig:PoissonG_ErrorAdam_3}
    \end{subfigure}%
    \begin{subfigure}{.4\textwidth}
        \centering
        \includegraphics[width=\linewidth]{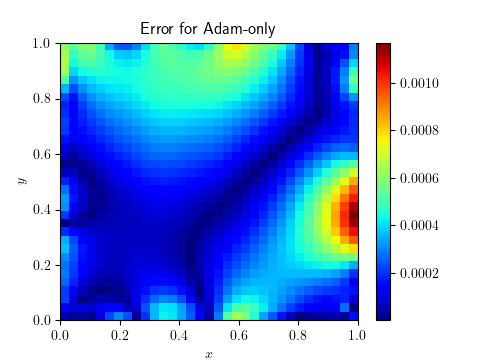}
        \includegraphics[width=\linewidth]{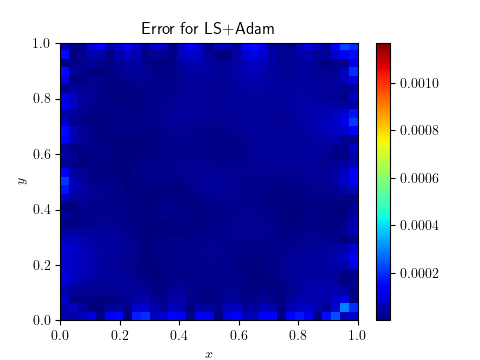}
        \caption{Unsupervised }\label{fig:PoissonG_ErrorLSAdam_2}
    \end{subfigure}
    \caption{\textit{Poisson equation with a variable BC:\@ Adam-only vs.
    LS+Adam, supervised and unsupervised.}
    (a)~Test data: input boundary condition $\tilde{g}$.
    (b)~Reference solution.
    (c), (d)~Top: absolute error of the DeepONet solution with Adam-only at $10^5$ WU\@.
    Bottom: absolute error of the DeepONet solution with LS+Adam at $10^4$ WU\@.
    For unsupervised cases, $\tilde{g}$ and $u$ are scaled down by a factor of $0.1$.
    }\label{fig:PoissonG_TestData}
\end{figure}

\Cref{fig:PoissonG_TestData} illustrates the solution errors of DeepONet
with Adam-only and LS+Adam for a test data.
In the supervised learning, the $L^2$ error of Adam-only at 100,000 WU is $\num{4.50e-3}$,
while the $L^2$ error of LS+Adam at 10,000 WU is $\num{1.34e-3}$.
In the unsupervised learning, the $L^2$ error of Adam-only at 100,000 WU is $\num{3.28e-4}$,
while the $L^2$ error of LS+Adam at 10,000 WU is $\num{4.46e-5}$.
In both supervised and unsupervised cases,
the errors are primarily concentrated near the boundary, while the interior errors
tend to be relatively very small. This tendency is more apparent in the LS+Adam cases.
To improve the imbalance between the near-boundary errors and interior errors,
an augmented Lagrangian term can be used for the BC constraint~\cite{Jang2024}.

\subsubsection{Poisson equation with a variable source}\label{subsubsec443}
In this section, we learn a solution operator from the source $f$ to $u$ via DeepONet
with unsupervised learning when $\kappa=1$ and $g=0$.
The 2D input source $f$ is generated from a GP with zero mean and a 2D squared exponential
covariance kernel \cref{eq:SqExpKer2D} with parameters $l_x=l_y=0.2$ and $\sigma^2 = 1$.
We take $33\times33$ equidistant grid points in ${[0,1]}^2$ for the
discretization of the input function and use $33\times33$ equidistant grid points of
${[0,1]}^2$ as the evaluation points of the output function.
The CNN of the branch network consists of three layers with $3\times 3$, $2\times 2$, and
$2\times 2$ kernels each with $2 \times 2$ strides. Since the channel sizes are $[1,16,32,64]$,
the output of the CNN is a $4\times 4$ image of $64$ channels.
Note that as mentioned in \cref{Tab:DON_config_Unsup},
we use a training that starts with $\lambda = 10^{-9}$ at 100 WU and decreases to
$\lambda = 10^{-14}$ at 1,000 WU\@.

\Cref{fig:UnSupLoss1}\nolinebreak(b)
shows that LS+Adam outperforms Adam-only in terms of training loss decay and
model performance for unsupervised learning with 2D sources as input
functions\@. The LS+Adam case reached the same level of
mean relative $L^2$ error around 10,000 WU, while the Adam-only case required 100,000 WUs.

\begin{figure}
    \centering
    \begin{subfigure}{.4\textwidth}
        \centering
        \includegraphics[width=\linewidth]{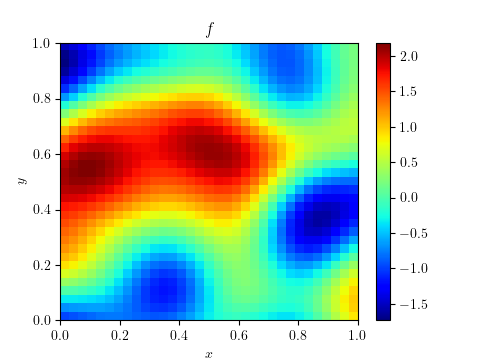}
        \caption{}\label{fig:PoissonF_Input}
    \end{subfigure}%
    \begin{subfigure}{.4\textwidth}
        \centering
        \includegraphics[width=\linewidth]{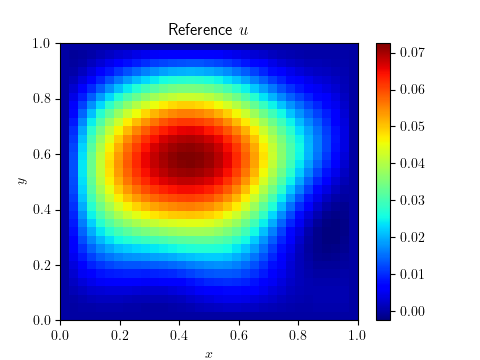}
        \caption{}\label{fig:PoissonF_Exact}
    \end{subfigure}
    \begin{subfigure}{.4\textwidth}
        \centering
        \includegraphics[width=\linewidth]{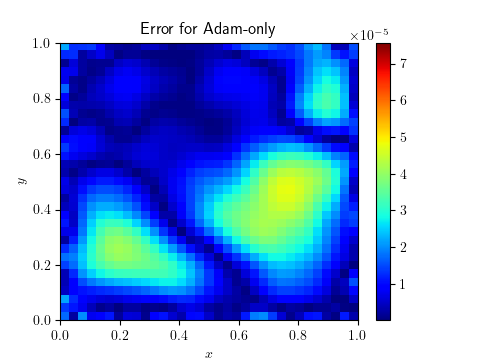}
        \caption{}\label{fig:PoissonF_ErrorAdam_2}
    \end{subfigure}%
    \begin{subfigure}{.4\textwidth}
        \centering
        \includegraphics[width=\linewidth]{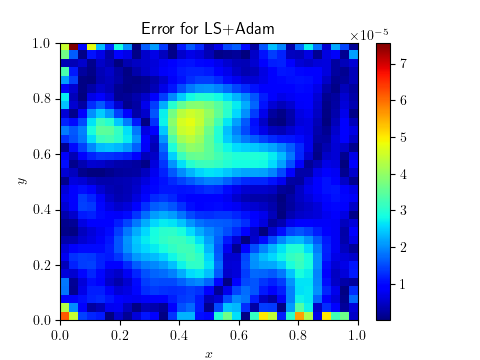}
        \caption{}\label{fig:PoissonF_ErrorLSAdam_2}
    \end{subfigure}
    \caption{\textit{Poisson equation with a variable source:\@ Adam-only vs.
    LS+Adam.}
    (a)~Test data: input source $f(x,y)$.
    (b)~Reference solution.
    (c) Absolute error of the DeepONet solution with Adam-only at $10^5$ WU\@.
    (d) Absolute error of the DeepONet solution with LS+Adam at $10^4$ WU\@.
    }\label{fig:PoissonF_TestData}
\end{figure}

\Cref{fig:PoissonF_TestData} illustrates the solution errors of DeepONet
with Adam-only and LS+Adam for a test data.
The $L^2$ error of Adam-only at 100,000 WU is $\num{1.86e-5}$,
while the $L^2$ error of LS+Adam at 10,000 WU is $\num{1.69e-5}$.
Unlike the case of a variable boundary in \cref{subsubsec442},
interior errors are prominent because we impose a variable source term that affects the
interior solution.

\section{Conclusion}\label{sec5}

In this paper, we proposed effective methods to accelerate the training of DeepONets.
By interpreting DeepONet with a linear last layer for the branch network as a linear
combination of results of the last hidden layer, whose coefficients are the last layer
parameters, we can optimize the last layer parameters using the least squares method.
This can be generalized to the sum of $L^2$ loss terms with linear operators and a
regularization term for the last layer parameters.
However, due to the network structure that includes the inner product and the large training
data size, forming and solving the least squares systems directly requires tremendous time and
memory.

To overcome these challenges, we assume that the data can be decomposed into branch and trunk
datasets, with each linear operator acting only on the output of the trunk network.
These assumptions allow us to factor the large LS system into two independent smaller systems
from the branch and trunk networks. Here, the LS system solution satisfies a
specific type of matrix equation, where the solution can be represented in closed form
via spectral decomposition.
Finally, the hybrid Least Squares/Gradient Descent method for DeepONets alternates between LS
and GD steps, where the LS step optimizes the last layer parameters by the above method and
the GD step optimizes the remaining hidden layer parameters using a GD-type optimizer.

Numerical experiments using various PDE examples
show that the LS+Adam method is highly effective in accelerating training in terms of
training loss reduction and relative errors on validation data compared to the traditional
training using Adam-only.

\bibliographystyle{siam}
\bibliography{references}

\begin{thebibliography}{10}

\bibitem{Augustine2024}
{\sc M.~T. Augustine}, {\em A survey on universal approximation theorems}, arXiv preprint arXiv:2407.12895,  (2024).

\bibitem{Barron1993}
{\sc A.~R. Barron}, {\em Universal approximation bounds for superpositions of a sigmoidal function}, IEEE Transactions on Information Theory, 39 (1993), pp.~930--945.

\bibitem{Bartels1972}
{\sc R.~H. Bartels and G.~W. Stewart}, {\em Algorithm 432 {[C2]}: solution of the matrix equation {AX+XB=C} {[F4]}}, Communications of the ACM, 15 (1972), pp.~820--826.

\bibitem{Baydin2018}
{\sc A.~G. Baydin, B.~A. Pearlmutter, A.~A. Radul, and J.~M. Siskind}, {\em Automatic differentiation in machine learning: a survey}, Journal of Machine Learning Research, 18 (2018), pp.~1--43.

\bibitem{jax2018}
{\sc J.~Bradbury, R.~Frostig, P.~Hawkins, M.~J. Johnson, C.~Leary, D.~Maclaurin, G.~Necula, A.~Paszke, J.~Vander{P}las, S.~Wanderman-{M}ilne, and Q.~Zhang}, {\em {JAX}: composable transformations of {P}ython+{N}um{P}y programs}, 2018.

\bibitem{Chen1995}
{\sc T.~Chen and H.~Chen}, {\em Universal approximation to nonlinear operators by neural networks with arbitrary activation functions and its application to dynamical systems}, IEEE Transactions on Neural Networks, 6 (1995), pp.~911--917.

\bibitem{Chu1987}
{\sc K.-w.~E. Chu}, {\em The solution of the matrix equations {AXB-CXD=E} and {(YA-DZ, YC-BZ)=(E, F)}}, Linear Algebra and its Applications, 93 (1987), pp.~93--105.

\bibitem{Cybenko1989}
{\sc G.~Cybenko}, {\em Approximation by superpositions of a sigmoidal function}, Mathematics of Control, Signals and Systems, 2 (1989), pp.~303--314.

\bibitem{Cyr2020}
{\sc E.~C. Cyr, M.~A. Gulian, R.~G. Patel, M.~Perego, and N.~A. Trask}, {\em Robust training and initialization of deep neural networks: An adaptive basis viewpoint}, in Mathematical and Scientific Machine Learning, PMLR, 2020, pp.~512--536.

\bibitem{Dettmers2019}
{\sc T.~Dettmers and L.~Zettlemoyer}, {\em Sparse networks from scratch: Faster training without losing performance}, {arXiv preprint arXiv:1907.04840},  (2019).

\bibitem{Golub1999}
{\sc G.~H. Golub, P.~C. Hansen, and D.~P. O'Leary}, {\em Tikhonov regularization and total least squares}, SIAM Journal on Matrix Analysis and Applications, 21 (1999), pp.~185--194.

\bibitem{He2015}
{\sc K.~He, X.~Zhang, S.~Ren, and J.~Sun}, {\em Delving deep into rectifiers: Surpassing human-level performance on imagenet classification}, in Proceedings of the IEEE International Conference on Computer Vision, 2015, pp.~1026--1034.

\bibitem{Heinecke2020}
{\sc A.~Heinecke, J.~Ho, and W.-L. Hwang}, {\em Refinement and universal approximation via sparsely connected {ReLU} convolution nets}, IEEE Signal Processing Letters, 27 (2020), pp.~1175--1179.

\bibitem{Hornik1989}
{\sc K.~Hornik, M.~Stinchcombe, and H.~White}, {\em Multilayer feedforward networks are universal approximators}, Neural Networks, 2 (1989), pp.~359--366.

\bibitem{Jang2024}
{\sc D.-K. Jang, K.~Kim, and H.~H. Kim}, {\em Partitioned neural network approximation for partial differential equations enhanced with {L}agrange multipliers and localized loss functions}, Computer Methods in Applied Mechanics and Engineering, 429 (2024), p.~117168.

\bibitem{Jin2022}
{\sc P.~Jin, S.~Meng, and L.~Lu}, {\em {MIONET}: Learning multiple-input operators via tensor product}, SIAM Journal on Scientific Computing, 44 (2022), pp.~A3490--A3514.

\bibitem{Kingma2017}
{\sc D.~P. Kingma}, {\em Adam: A method for stochastic optimization}, arXiv preprint arXiv:1412.6980,  (2014).

\bibitem{Lee2022}
{\sc Y.~Lee, J.~Park, and C.-O. Lee}, {\em Two-level group convolution}, Neural Networks, 154 (2022), pp.~323--332.

\bibitem{Lee2024}
\leavevmode\vrule height 2pt depth -1.6pt width 23pt, {\em Parareal neural networks emulating a parallel-in-time algorithm.}, IEEE Transactions on Neural Networks and Learning Systems, 35 (2024), pp.~6353--6364.

\bibitem{Li2020}
{\sc Z.~Li, N.~Kovachki, K.~Azizzadenesheli, B.~Liu, K.~Bhattacharya, A.~Stuart, and A.~Anandkumar}, {\em Fourier neural operator for parametric partial differential equations}, {arXiv preprint arXiv:2010.08895},  (2020).

\bibitem{Li2020neural}
\leavevmode\vrule height 2pt depth -1.6pt width 23pt, {\em Neural operator: Graph kernel network for partial differential equations}, {arXiv preprint arXiv:2003.03485},  (2020).

\bibitem{Lin2018}
{\sc H.~Lin and S.~Jegelka}, {\em Resnet with one-neuron hidden layers is a universal approximator}, Advances in Neural Information Processing Systems, 31 (2018).

\bibitem{Liu1989}
{\sc D.~C. Liu and J.~Nocedal}, {\em On the limited memory {BFGS} method for large scale optimization}, Mathematical Programming, 45 (1989), pp.~503--528.

\bibitem{Lu2021}
{\sc L.~Lu, P.~Jin, G.~Pang, Z.~Zhang, and G.~E. Karniadakis}, {\em Learning nonlinear operators via {DeepONet} based on the universal approximation theorem of operators}, Nature Machine Intelligence, 3 (2021), pp.~218--229.

\bibitem{Lu2022}
{\sc L.~Lu, X.~Meng, S.~Cai, Z.~Mao, S.~Goswami, Z.~Zhang, and G.~E. Karniadakis}, {\em A comprehensive and fair comparison of two neural operators (with practical extensions) based on fair data}, Computer Methods in Applied Mechanics and Engineering, 393 (2022), p.~114778.

\bibitem{Lu2017}
{\sc Z.~Lu, H.~Pu, F.~Wang, Z.~Hu, and L.~Wang}, {\em The expressive power of neural networks: A view from the width}, Advances in Neural Information Processing Systems, 30 (2017).

\bibitem{Lu2024}
{\sc Z.~Lu, Y.~Zhou, Y.~Zhang, X.~Hu, Q.~Zhao, and X.~Hu}, {\em A fast general thermal simulation model based on multi-branch physics-informed deep operator neural network}, Physics of Fluids, 36 (2024), p.~037142.

\bibitem{Magnus1979}
{\sc J.~R. Magnus and H.~Neudecker}, {\em {The Commutation Matrix: Some Properties and Applications}}, The Annals of Statistics, 7 (1979), pp.~381 -- 394.

\bibitem{Neudecker1968}
{\sc H.~Neudecker}, {\em The {K}ronecker matrix product and some of its applications in econometrics}, Statistica Neerlandica, 22 (1968), pp.~69--82.

\bibitem{Raissi2019}
{\sc M.~Raissi, P.~Perdikaris, and G.~E. Karniadakis}, {\em Physics-informed neural networks: A deep learning framework for solving forward and inverse problems involving nonlinear partial differential equations}, Journal of Computational Physics, 378 (2019), pp.~686--707.

\bibitem{Ramachandran2017}
{\sc P.~Ramachandran, B.~Zoph, and Q.~V. Le}, {\em Searching for activation functions}, arXiv preprint arXiv:1710.05941,  (2017).

\bibitem{Salimans2016}
{\sc T.~Salimans and D.~P. Kingma}, {\em Weight normalization: A simple reparameterization to accelerate training of deep neural networks}, Advances in Neural Information Processing Systems, 29 (2016).

\bibitem{Son2025}
{\sc H.~Son}, {\em {ELM-DeepONets}: Backpropagation-free training of deep operator networks via extreme learning machines}, IEEE Access, 13 (2025), pp.~86927--86934.

\bibitem{Tikhonov1963}
{\sc A.~N. Tikhonov}, {\em On the solution of ill-posed problems and the method of regularization}, in Doklady Akademii Nauk, vol.~151, Russian Academy of Sciences, 1963, pp.~501--504.

\bibitem{Wang2021}
{\sc S.~Wang, H.~Wang, and P.~Perdikaris}, {\em Learning the solution operator of parametric partial differential equations with physics-informed {D}eep{ON}ets}, Science {A}dvances, 7 (2021), p.~eabi8605.

\bibitem{Zheng2016}
{\sc S.~Zheng, A.~Vishnu, and C.~Ding}, {\em Accelerating deep learning with shrinkage and recall}, in {2016 IEEE 22nd International Conference on Parallel and Distributed Systems (ICPADS)}, IEEE, 2016, pp.~963--970.

\bibitem{Zhou2020}
{\sc D.-X. Zhou}, {\em Universality of deep convolutional neural networks}, Applied and Computational Harmonic Analysis, 48 (2020), pp.~787--794.

\end{thebibliography}

\end{document}